\documentclass[10pt,twocolumn]{article}
\usepackage{graphicx}
\usepackage{amsmath,amssymb,times,cite}

\DeclareMathOperator{\diag}{diag}
\DeclareMathOperator{\tr}{tr}

\newcommand{\bm}[1]{\mathbf{#1}} 

\newcommand\T{{\mathpalette\raiseT\intercal}}
\newcommand\raiseT[2]{%
\setbox0\hbox{$#1{#2}$}\raise\dp0\box0}

\usepackage{booktabs}
\title{\Large\textbf{Anisotropic Graph Convolutional Network for Semi-supervised Learning}}
\author{Mahsa Mesgaran and A. Ben Hamza\\
Concordia Institute for Information Systems Engineering\\
Concordia University, Montreal, QC, Canada
}
\date{}

\begin{document}
\maketitle

\begin{abstract}
Graph convolutional networks learn effective node embeddings that have proven to be useful in achieving high-accuracy prediction results in semi-supervised learning tasks, such as node classification. However, these networks suffer from the issue of over-smoothing and shrinking effect of the graph due in large part to the fact that they diffuse features across the edges of the graph using a linear Laplacian flow. This limitation is especially problematic for the task of node classification, where the goal is to predict the label associated with a graph node. To address this issue, we propose an anisotropic graph convolutional network for semi-supervised node classification by introducing a nonlinear function that captures informative features from nodes, while preventing oversmoothing. The proposed framework is largely motivated by the good performance of anisotropic diffusion in image and geometry processing, and learns nonlinear representations based on local graph structure and node features. The effectiveness of our approach is demonstrated on three citation networks and two image datasets, achieving better or comparable classification accuracy results compared to the standard baseline methods.
\end{abstract}


\section{Introduction}
Graphs are ubiquitous in a wide array of application domains, ranging from social networks~\cite{Bhagat:11,FHuang:17,LXu:19} and transportation systems~\cite{SGuo:19} and cyber-security~\cite{SWang:19} to brain networks~\cite{Ktena:18} graph signal processing~\cite{Ortega:18}, and video analysis~\cite{YChen:19,QPeng:19,JGao:20}. They provide a flexible way to inherently represent real-world entities as a set of nodes and their interactions as a set of links/edges. This interconnection of entities and their pairwise relationships forms a graph structure when visualized.

With the prevalence and increasing proliferation of graph-structured data in real-world applications, there has been a surge of interest in developing efficient representations of graphs. Network embedding has recently emerged as a powerful paradigm for representing and analyzing graph-structured data~\cite{Weston:08,WYang:16,Goyal:18,LLu:20}. The idea is to learn low-dimensional embedding vectors, such that both structural and semantic information are captured. These learned embeddings can then be used as input to various machine learning algorithms for downstream tasks, such as link prediction, visualization, recommendation, community detection, and node classification. The latter task is the focus of this paper. The objective of node classification is to predict the most probable labels of nodes in a graph~\cite{Hamilton:17}. In a social network, for instance, we want to predict user labels such as their interest, beliefs or other characteristics~\cite{Bhagat:11}, while in a citation network, we want to classify documents based on their topics.

There is a sizable body of literature on network embedding that has centered around the use of random walks and neural language models to learn effective low-dimensional embedding vectors of graph nodes~\cite{Perozzi:14,Grover:16,Hamilton:17}. Perozzi \textit{et al.}~\cite{Perozzi:14} introduce DeepWalk, a deep learning based framework that learns latent representations of nodes in a graph by leveraging local information obtained from truncated random walks. Each random walk is treated as a sentence that is fed into the skip-gram language model~\cite{Mikolov:13}, which maximizes the co-occurrence probability among the words that appear within a window in a sentence. Another popular approach that also uses random walks is node2vec~\cite{Grover:16}, a semi-supervised algorithm for feature learning in graphs that can be regarded as a generalization of DeepWalk. While DeepWalk performs a uniform random walk, node2vec uses a second-order random walk approach to generate network neighborhoods for nodes via breadth-first and depth-first sampling strategies. However, node2vec involves a number of parameters that require fine-tuning for each dataset and each task.

In recent years, the advent of deep learning has sparked groundswell of interest in the adoption of graph neural networks (GNNs) for learning latent representations of graphs~\cite{Li:16,Bresson:18,Chenyi:18,Hongyang:18,Wenbing:18,Defferrard:16,Kipf:17,WLHamilton:17,Velickovic:17,KXu:19}. A plethora of GNNs is based on convolutional neural networks (CNNs) and network embedding. Defferrard \textit{et al.}~\cite{Defferrard:16} introduce the Chebyshev network (ChebyNet), an efficient spectral-domain graph convolutional neural network that uses recursive Chebyshev polynomial spectral filters to avoid explicit computation of the Laplacian eigenvectors. These filters are localized in space, and the learned weights can be shared across different locations in a graph. An efficient variant of GNNs is graph convolutional networks (GCNs)~\cite{Kipf:17}, which is an upsurging semi-supervised graph-based deep learning framework that uses an efficient layer-wise propagation rule based on a first-order approximation of spectral graph convolutions. Hamilton \textit{et al.}~\cite{WLHamilton:17} propose GraphSAGE, a general inductive framework that generates embeddings by sampling and aggregating features from the local neighborhood of a graph node. Veli\u{c}kovi\'{c} \textit{et al.}~\cite{Velickovic:17} present the graph attention network, which is a graph-based neural network architecture that uses an attention mechanism to assign self-attention scores to neighboring node embeddings. These scores indicate the importance of graph nodes to their corresponding neighbors on the feature aggregation process. Xu \textit{et al.}~\cite{KXu:19} present theoretical foundations for analyzing the expressive power of GNNs in an effort to capture different graph structures, and develop a graph isomorphism network whose goal is to map isomorphic graphs to the same representation and non-isomorphic ones to different representations.

While graph convolutional networks have achieved state-of-the-art performance on semi-supervised node classification tasks, they tend, however, to oversmooth the learned feature embeddings of graph nodes~\cite{Li:18}. This is due largely to the fact that the graph convolution of the GCN model is a special form of graph Laplacian smoothing, which repeatedly and simultaneously adjusts the location of each graph node to the weighted average (i.e. geometric center) of its neighboring nodes. This averaging process causes an oversmoothing effect on the graph, as it reduces the high-frequency graph information and tends to flatten the graph. Moreover, oversmoothing causes features at nodes within each connected component to converge to the same value. Hence, nodes from different classes may be predicted to have similar labels, resulting in misclassification errors. Another drawback of Laplacian smoothing is shrinkage of the graph, as repeated iterations of the smoothing process causes the shrinking effect.

In this paper, we propose an anisotropic graph convolutional network (AGCN), which adopts the concept of anisotropic diffusion, previously used in image and geometry processing tasks, to overcome the aforementioned issues. The idea behind our proposed model is to integrate a nonlinearity term into the graph convolution to make it non-linear and/or anisotropic, resulting in a feature-preserving graph neural network. The main contributions of this work can be summarized as follows:
\begin{itemize}
\item We introduce a novel anisotropic graph convolutional network for semi-supervised learning.
\item We learn efficient representations for node classification in an end-to-end fashion.
\item We demonstrate that AGCN can be integrated into existing graph-based convolutional networks for semi-supervised learning using both co-training and self-training.
\item Our extensive experimental results show competitive or superior performance of AGCN over standard baseline methods on several benchmark datasets.
\end{itemize}	

\medskip
The rest of this paper is organized as follows. In Section 2, we review important relevant work. In Section 3, we present the problem formulation and propose an anisotropic graph convolutional network architecture for semi-supervised learning. We discuss in detail the main components of the proposed framework and analyze the model complexity. In Section 4, we present experimental results to demonstrate the competitive performance of our approach on five standard benchmark datasets, including three citations networks and two image datasets. Finally, we conclude in Section 5 and point out future work directions.

\section{Related Work}
The basic goal of node classification is to predict the most probable labels of nodes in a graph. Graph convolutional networks (GCNs) have recently become the de facto model for semi-supervised node classification~\cite{Kipf:17}. GCN uses an efficient layer-wise propagation rule, which is based on a first-order approximation of spectral graph convolutions. The feature vector of each graph node is updated by essentially applying a weighted sum of the features of its neighboring nodes. Monti \textit{et al.}~\cite{Monti:17} present a mixture of networks (MoNet) model, a spatial-domain graph convolutional neural network that employs a mixture of Gaussian kernels with learnable parameters to model the weight function of pseudo-coordinates, which are associated to the neighboring nodes of each graph node. Liao \textit{et al.}~\cite{Liao:19} propose the Lanczos network (LanczosNet), which employs the Lanczos algorithm to construct low-rank approximations of the graph Laplacian in order to facilitate efficient computations of matrix powers. Veli\u{c}kovi\'{c} \textit{et al.}~\cite{Velickovic:19} present deep graph infomax, an unsupervised graph representation learning approach, which relies on training an encoder model to maximize the mutual information between local and global representations in graphs. Xu \textit{et al.} ~\cite{Bingbing:19} introduce a graph wavelet neural network, which is a GCN-based architecture that uses spectral graph wavelets in lieu of graph Fourier bases to define a graph convolution. Despite the fact that spectral graph wavelets can yield localization of graph signals in both spatial and spectral domains, they require explicit computation of the Laplacian eigenbasis, leading to a high computational complexity, especially for large graphs. In order to avoid this issue, recursive Chebyshev polynomial spectral filters can be employed.

While GCNs have shown great promise, achieving state-of-the-art performance on semi-supervised node classification, they are prone to oversmoothing the node features. In fact, the neighborhood aggregation scheme (i.e. graph convolution) of GCN is tantamount to applying Laplacian smoothing~\cite{Li:18}, which replaces each graph node with the average of its immediate neighbors. Therefore, repeated application of GCN yields smoother and smoother versions of the initial node features as the number of the network's layers increases. As a result, the node features in deeper layers will eventually converge to the same value, and hence become too similar across different classes. Wu \textit{et al.}~\cite{FWu:19} introduce a simple graph convolution by removing the nonlinear transition functions between the layers of graph convolutional networks and collapsing the resulting function into a single linear transformation via the powers of the normalized adjacency matrix with added self-loops for all graph nodes. However, this simple graph convolution acts as a low-pass filter, which attenuates all but the zero frequency, causing oversmoothing. Recently, significant strides have been made toward remedying the issue of oversmoothing in GCNs~\cite{Keyulu:18,Lingxiao:20}. Xu \textit{et al.}\cite{Keyulu:18} propose jumping knowledge networks, which employ dense skip connections to connect each layer of the network with the last layer to preserve the locality of node representations in order circumvent oversmoothing. More recently, a normalization layer, which helps avoid oversmoothing by preventing learned representations of distant nodes from becoming indistinguishable, has been proposed in~\cite{Lingxiao:20}. This normalization layer is performed on intermediate layers during training, and the aim is to apply smoothing over nodes within the same cluster while avoiding smoothing over nodes from different clusters. While these approaches have shown slightly improved results using deeper GCNs, the issue of oversmoothing still remains a daunting task, as performance gains do not usually reflect the benefits of increasing the network depth.

\section{Method}
In this section, we describe the problem statement and introduce an anisotropic graph convolutional network for semi-supervised node classification. In particular, we examine the main building blocks of the proposed network architecture and analyze the complexity of the model. We also show that our proposed aggregation scheme seamlessly incorporates both the graph structure and the node features without sacrificing performance in an effort to alleviate oversmoothing of the learned node representations.

\subsection{Problem Formulation}
Let $\mathbb{G}=(\mathcal{V},\mathcal{E})$ be a graph, where $\mathcal{V}=\{1,\ldots,N\}$ is the set of $N$ nodes and $\mathcal{E}\subseteq \mathcal{V}\times\mathcal{V}$ is the set of edges. We denote by $\bm{A}=(\bm{A}_{ij})$ an $N\times N$ adjacency matrix (binary or real-valued) whose $(i,j)$-th entry $\bm{A}_{ij}$ is equal to the weight of the edge between neighboring nodes $i$ and $j$, and 0 otherwise. We also denote by $\bm{X}=(\bm{x}_{1},...,\bm{x}_{N})^{\T}$ an $N\times F$ feature matrix of node attributes, where $\bm{x}_{i}$ is an $F$-dimensional row vector for node $i$.

Learning latent representations of nodes in a graph aims at encoding the graph structure into low-dimensional embeddings, such that both structural and semantic information are captured. More precisely, the purpose of network/graph embedding is to learn a mapping $\varphi: \mathcal{V}\to\mathbb{R}^{P}$ that maps each node $i$ to a $P$-dimensional vector $\bm{z}_i$, where $P\ll N$. These learned node embeddings can then be used as input to learning algorithms for downstream tasks, such as node classification.

Given the labels of a subset of the graph nodes (or their corresponding final output embeddings), the objective of semi-supervised learning is to predict the unknown labels of the other nodes. More specifically, let $\mathcal{D}_{K}=\{(\bm{z}_i,y_i)\}_{i=1}^{K}$ be the set of labeled final output node embeddings $\bm{z}_i\in\mathbb{R}^{P}$ with associated known labels $y_i\in\mathcal{Y}_K$, and $\mathcal{D}_{U}=\{\bm{z}_i\}_{i=K+1}^{K+U}$ be the set of unlabeled final output node embeddings, where $K+U=N$. Then, the problem of semi-supervised node classification is to learn a classifier $f: \mathcal{V}\to\mathcal{Y}_K$. That is, the goal is to predict the labels of the set $\mathcal{D}_{U}$.

It is important to note that for multi-class classification problems, the label of each node $i$ (or its final output embedding $\bm{z}_i$) in the labeled set $\mathcal{D}_{K}$ can be represented as a $C$-dimensional one-hot vector $\bm{y}_{i} \in \{0, 1\}^{C}$, where $C$ is the number of classes.

\subsection{Proposed Approach}
Graph convolutional networks learn a new feature representation for each node such that nodes with the same labels have similar features~\cite{Kipf:17}. Given a graph $\mathbb{G}=(\mathcal{V},\mathcal{E})$ with adjacency matrix $\bm{A}\in\mathbb{R}^{N\times N}$ and feature matrix $\bm{X}\in\mathbb{R}^{N\times F}$, the layer-wise feature diffusion rule of an $L$-layer GCN is given by
\begin{equation}
\bm{S}^{(\ell)}=\tilde{\bm{D}}^{-\frac{1}{2}}\tilde{\bm{A}}\tilde{\bm{D}}^{-\frac{1}{2}}\bm{H}^{(\ell)},
\quad \ell=0,\dots,L-1,
\label{Eq:GCNdiff}
\end{equation}
where $\tilde{\bm{A}}=\bm{A}+\bm{I}_{N}$ is the adjacency matrix with self-added loops, $\bm{I}_{N}$ is the identity matrix, $\tilde{\bm{D}}=\diag(\tilde{d}_{i})$ is the diagonal degree matrix whose $i$-th diagonal entry is the degree of node $i$ with added self-loops, and $\bm{H}^{(\ell)}\in\mathbb{R}^{N\times F_{\ell}}$ is the input feature matrix of the $\ell$-th layer with $F_{\ell}$ feature maps. The input of the first layer is the original feature matrix $\bm{H}^{(0)}=\bm{X}$.

Using the feature diffusion rule of GCN is tantamount to applying a weighted sum of the features of neighboring nodes normalized by their degrees, which essentially performs Laplacian smoothing on the graph~\cite{Li:18,FWu:19}. In other words, the smooth feature matrix $\bm{S}^{(\ell)}$ is obtained by applying Laplacian smoothing to the input feature matrix at the $\ell$-th layer. Intuitively, the Laplacian flow repeatedly and simultaneously adjusts the location of each graph node to the geometric center of its neighboring nodes. Although the Laplacian smoothing flow is simple and fast, it produces, however, the shrinking effect and an oversmoothing result.

Motivated by the good performance of anisotropic diffusion in image and mesh denoising~\cite{Weickert:98,Black:98,Zhang:07}, and in an effort to tackle the issues of oversmoothing and shrinking effect of GCN, we propose an anisotropic graph convolutional network (AGCN) for semi-supervised node classification by incorporating a nonlinear smoothness term into the GCN feature diffusion rule. This nonlinearity term, which quantifies the dissimilarity between learned node embeddings, plays a pivotal role in preventing these learned representations from becoming increasingly similar, and hence alleviates the issue of oversmoothing. In addition, it tackles the shrinking effect by precluding the learned node representations from converging to the same value.

\medskip
\noindent{\textbf{Anisotropic feature diffusion.}}\quad We define a layer-wise anisotropic feature diffusion rule for node features in the $\ell$-th layer as follows:
\begin{equation}
\bm{G}^{(\ell)} =\Bigl(1-\exp\bigl(-\beta\tr^{2}(\bm{H}^{(\ell)\T}\tilde{\bm{L}}\bm{H}^{(\ell)})\bigr)\Bigr)
\tilde{\bm{D}}^{-\frac{1}{2}}\tilde{\bm{A}}\tilde{\bm{D}}^{-\frac{1}{2}}\bm{H}^{(\ell)},
\end{equation}
where $\beta$ is an nonnegative hyper-parameter that is often fine-tuned via grid search, and $\tr(\bm{H}^{(\ell)\T}\tilde{\bm{L}}\bm{H}^{(\ell)})$ is a Laplacian smoothness term given by
\begin{equation}
\tr(\bm{H}^{(\ell)\T}\tilde{\bm{L}}\bm{H}^{(\ell)}) =\frac{1}{2}\sum_{i,j=1}^{N}\tilde{\bm{A}}_{ij}\Vert\bm{h}_{i}^{(\ell)}-\bm{h}_{j}^{(\ell)}\Vert^{2},
\end{equation}
with $\bm{H}^{(\ell)}=(\bm{h}_{1}^{(\ell)},\ldots,\bm{h}_{N}^{(\ell)})^{\T}$; $\bm{h}_{i}^{(\ell)}$ is an $F_\ell$-dimensional hidden representation (embedding) vector of the $i$-th node at the $\ell$-th layer, $\tr(\cdot)$ denotes the trace operator, $\Vert\cdot\Vert$ denotes the 2-norm, and $\tilde{\bm{L}}=\tilde{\bm{D}}-\tilde{\bm{A}}$ is an $N\times N$ Laplacian matrix.

For each pair of similar embeddings at the $\ell$-th layer, the Laplacian smoothness term enforces their predictions to be close to each other. The strength of this smoothness is determined by the weight of the edge between neighboring nodes, meaning that connected nodes will have similar predictions.

The Laplacian smoothness term plays a crucial role not only in explicitly taking into consideration the correlation between embeddings, but also in preserving the locality of nodes to be embedded. In other words, two nodes or their attributes $\bm{x}_i$ and $\bm{x}_j$ that are close to each other in the original graph (i.e. adjacent nodes in $\mathcal{V}$) are encoded as embeddings $\bm{h}_{i}^{(\ell)}$ and $\bm{h}_{j}^{(\ell)}$ that are more likely to be close to each other in the embedding vector space. Such a locality-preserving property is of paramount importance in classification tasks.

The nonlinearity term $1-\exp\bigl(-\beta\tr^{2}(\bm{H}^{(\ell)\T}\tilde{\bm{L}}\bm{H}^{(\ell)})\bigr)$ can be regarded as an oversmooting ``stopping'' function. In fact, it only incurs a small penalty when similar nodes with a large smoothness strength $\tilde{\bm{A}}_{ij}$ have different learned embeddings. Hence, it reduces the oversmoothing effect on the learned graph features.

\begin{figure}[!htb]
\centering
\includegraphics[scale=0.54]{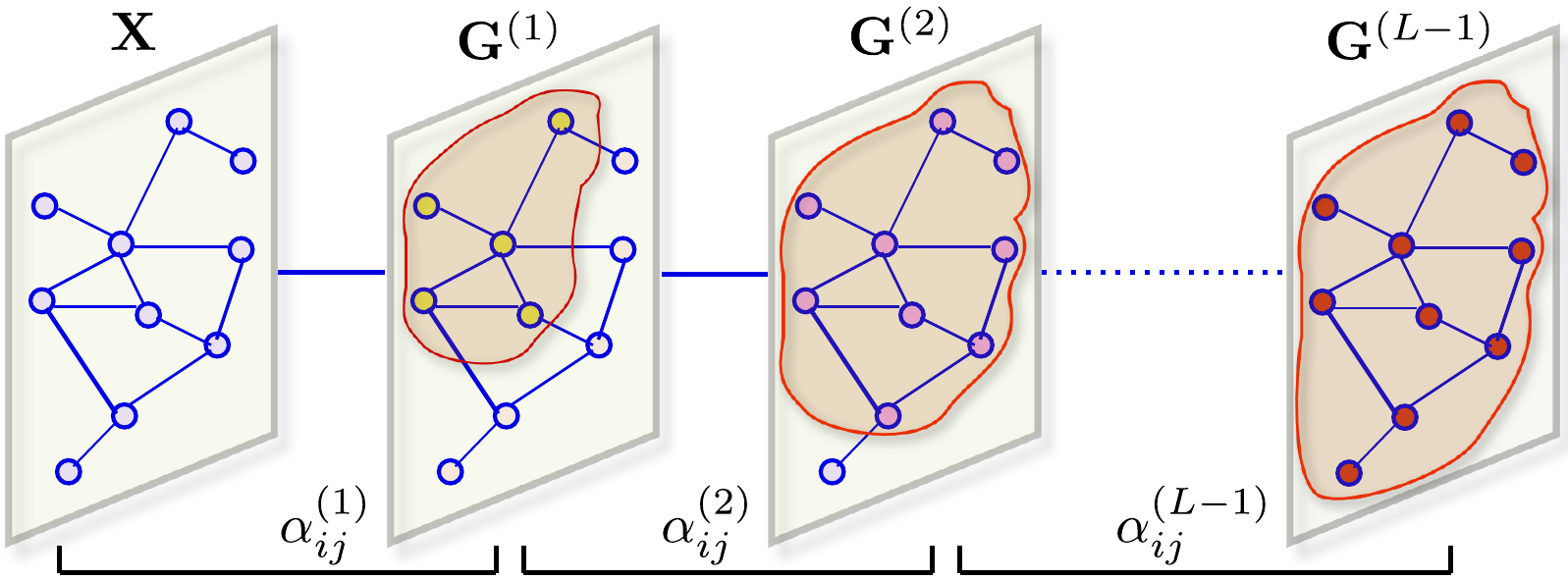}
\caption{Schematic layout of the anisotropic aggregation procedure.}
\label{Fig:AGCNscheme}
\end{figure}

\medskip
\noindent{\textbf{Anisotropic aggregation procedure.}}\quad The anisotropic feature diffusion rule can be written in vector form as follows:
\begin{equation}
\bm{g}_{i}^{(\ell)} = \sum_{j=1}^{N} \alpha_{ij}^{(\ell)}\bm{h}_{j}^{(\ell)}
\end{equation}
where $\bm{g}_{i}^{(\ell)}$ is the $i$-th row of $\bm{G}_{i}^{(\ell)}$, and $\alpha_{ij}^{(\ell)}$ is the layer-wise weight coefficient given by
\begin{equation}
\alpha_{ij}^{(\ell)}=\Bigl(1-\exp\bigl(-\beta\tr^{2}(\bm{H}^{(\ell)\T}\tilde{\bm{L}}\bm{H}^{(\ell)})\bigr)\Bigr)\frac{\tilde{\bm{A}}_{ij}}{\sqrt{\tilde{d}_{i}\tilde{d}_{j}}}
\end{equation}
which is equal to the nonlinearity term times the weight of the GCN neighborhood aggregation. The anisotropic aggregation scheme is illustrated in Figure~\ref{Fig:AGCNscheme}. Notice how features are propagated from 1-hop (i.e. immediate) neighbors to multi-hop (i.e. distant) neighbors as the number of layers increases.

Unlike the GCN weight which only takes into account the topological structure of the graph, our proposed anisotropic feature diffusion rule seamlessly leverages both the topological structure and nodal attributes for aggregating node representations. Also, it is important to note that compared to existing neighborhood aggregation approaches~\cite{Kipf:17,WLHamilton:17,Velickovic:17}, the weight coefficient $\alpha_{ij}^{(\ell)}$ of our aggregation scheme is layer-aware, and hence prevents the learned representations from becoming indistinguishable thanks to the synergy between the nonlinearity term that helps mitigate oversmoothing of the node features and the GCN weight that captures the graph structure.

As illustrated in Figure~\ref{Fig:AGCNscheme}, the proposed anisotropic feature diffusion rule follows a neighborhood aggregation or a message passing algorithm to learn a node representation by propagating representations of its immediate neighbors, where the latent representation of each node is initialized to the node's input features. The latent representation of each graph node at a given layer is defined as a weighted sum of its immediate neighbors' representations from the previous layer. As the number of layers increases, the node features are propagated to higher-order neighborhoods.

\medskip
\noindent{\textbf{Learning embeddings.}}\quad Given the anisotropically smooth feature matrix $\bm{G}^{(\ell)}$ at the $\ell$-th layer as an input, the output feature matrix $\bm{G}^{(\ell+1)}$ of our proposed AGCN model is obtained by applying the following layer-wise propagation rule:
\begin{equation}
\bm{G}^{(\ell+1)}=\sigma(\bm{G}^{(\ell)}\bm{W}^{(\ell)}),\quad \ell=0,\dots,L-1,
\label{Eq:AGCNprop}
\end{equation}
which is basically a node embedding transformation that projects the input $\bm{G}^{(\ell)}\in\mathbb{R}^{N\times F_{\ell}}$ into a trainable weight matrix $\bm{W}^{(\ell)}\in\mathbb{R}^{F_{\ell}\times F_{\ell+1}}$ with $F_{\ell +1}$ feature maps, followed by a point-wise non-linear activation function $\sigma(\cdot)$ such as $\text{ReLU}(\cdot)=\max(0,\cdot)$, assuming that $F_{\ell+1}\le F_{\ell}\ll N$. It is worth pointing out that after feature aggregation, performing the AGCN layer-wise propagation rule amounts to applying a multi-layer perceptron to the anisotropic feature matrix. In other words, a node latent representation at layer $\ell$ is transformed linearly via a learned weight matrix to produce the node latent representation at the next layer.

\medskip
\noindent{\textbf{Model prediction.}} The embedding $\bm{G}^{(L)}$ of the last layer of AGCN contains the final output node embeddings, and captures the neighborhood structural information of the graph within $L$ hops. This final node representation can be used as input for downstream tasks such as graph classification, clustering, visualization, link prediction, and node classification. Since the latter task is the focus of this paper, we apply a softmax classifier as follows:
\begin{equation}
\hat{\bm{Y}}=\text{softmax}(\bm{G}^{(L)}\bm{W}^{(L)}),
\end{equation}
where $\bm{W}^{(L)}\in\mathbb{R}^{F_{L}\times C}$ is a trainable weight matrix of the last layer, $C$ is the total number of classes, $\text{softmax}(\bm{x})=\exp(\bm{x})/\sum_{c=1}^{C}\exp(\bm{x}_{c})$ is an activation function applied row-wise, and $\hat{\bm{Y}}\in\mathbb{R}^{N\times C}$ is the matrix of predicted labels for graph nodes.

\medskip
\noindent{\textbf{Model complexity.}} For simplicity, we assume the feature dimensions are the same for all layers, i.e. $F_{\ell}=F$ for all $\ell$, with $F \ll N$. The time complexity of an $L$-layer AGCN is $\mathcal{O}(L\vert\mathcal{E}\vert F+LNF^2)$, where $\vert\mathcal{E}\vert$ denotes the number of graph edges. Note that multiplying the normalized adjacency matrix with an embedding costs $\mathcal{O}(\vert\mathcal{E}\vert F)$ in time, while multiplying an embedding with a weight matrix costs $\mathcal{O}(NF^2)$. Also, noting that $\tr(\bm{H}^{(\ell)\T}\tilde{\bm{L}}\bm{H}^{(\ell)})=\tr(\tilde{\bm{L}}\bm{H}^{(\ell)}\bm{H}^{(\ell)\T})$, it follows that computing this trace operator requires $NF^2$ scalar multiplications. Hence, the nonlinearity term of AGCN has complexity $\mathcal{O}(NF^2)$.

For memory complexity, an $L$-layer AGCN requires $\mathcal{O}(LNF+LF^2)$ in memory, where $\mathcal{O}(LNF)$ is for storing all embeddings and $\mathcal{O}(LF^2)$ is for storing all layer-wise weight matrices.

Therefore, the proposed AGCN model has the same time and memory complexity as GCN, while being effective at alleviating the issue of oversmoothing.

\medskip
\noindent{\textbf{Model training.}} For semi-supervised multi-class classification, the neural network weight parameters are learned by minimizing the cross-entropy loss function
\begin{equation}
\mathcal{L}=-\sum_{i\in \mathcal{Y}_{K}}\sum_{c=1}^{C} \bm{Y}_{ic} \log\hat{\bm{Y}}_{ic},
\end{equation}
over the set $\mathcal{Y}_{K}$ of all labeled nodes using gradient descent, where $\bm{Y}_{ic}$ is equal 1 if node $i$ belongs to class $c$, and 0 otherwise; and $\hat{\bm{Y}}_{ic}$ is the $(i,c)$-element of the matrix $\hat{\bm{Y}}$ from the softmax function, i.e. the probability that the network associates the $i$-th node with class $c$.

\section{Experiments}
In this section, we conduct extensive experiments to evaluate the performance of the proposed AGCN framework on several benchmark datasets and carry out a comprehensive comparison with several baseline methods. In all experiments, we consider a two-layer AGCN for semi-supervised node classification
\begin{equation}
\hat{\bm{Y}}=\text{softmax}\Bigl(\text{ReLU}\bigl(\bm{G}^{(0)}\bm{W}^{(0)}\bigr)\bm{W}^{(1)}\Bigr),
\label{Eq:twolayer}
\end{equation}
where $\bm{W}^{(0)}\in\mathbb{R}^{F\times F_{1}}$ is a trainable input-to-hidden weight matrix for a hidden layer with $F_1$ feature maps, $\bm{W}^{(1)}\in\mathbb{R}^{F_{1}\times C}$ is a trainable hidden-to-output weight matrix with $C$ denoting the number of classes, and $\bm{G}^{(0)}$ is an $N\times F$ matrix given by
\begin{equation}
\bm{G}^{(0)} =\Bigl(1-\exp\bigl(-\beta\tr^{2}(\bm{X}^\T\tilde{\bm{L}}\bm{X})\bigr)\Bigr)
\tilde{\bm{D}}^{-\frac{1}{2}}\tilde{\bm{A}}\tilde{\bm{D}}^{-\frac{1}{2}}\bm{X}.
\label{Eq:G0}
\end{equation}

\begin{figure}[!htp]
\setlength{\tabcolsep}{.25em}
\centering
\begin{tabular}{cc}
\includegraphics[scale=.34]{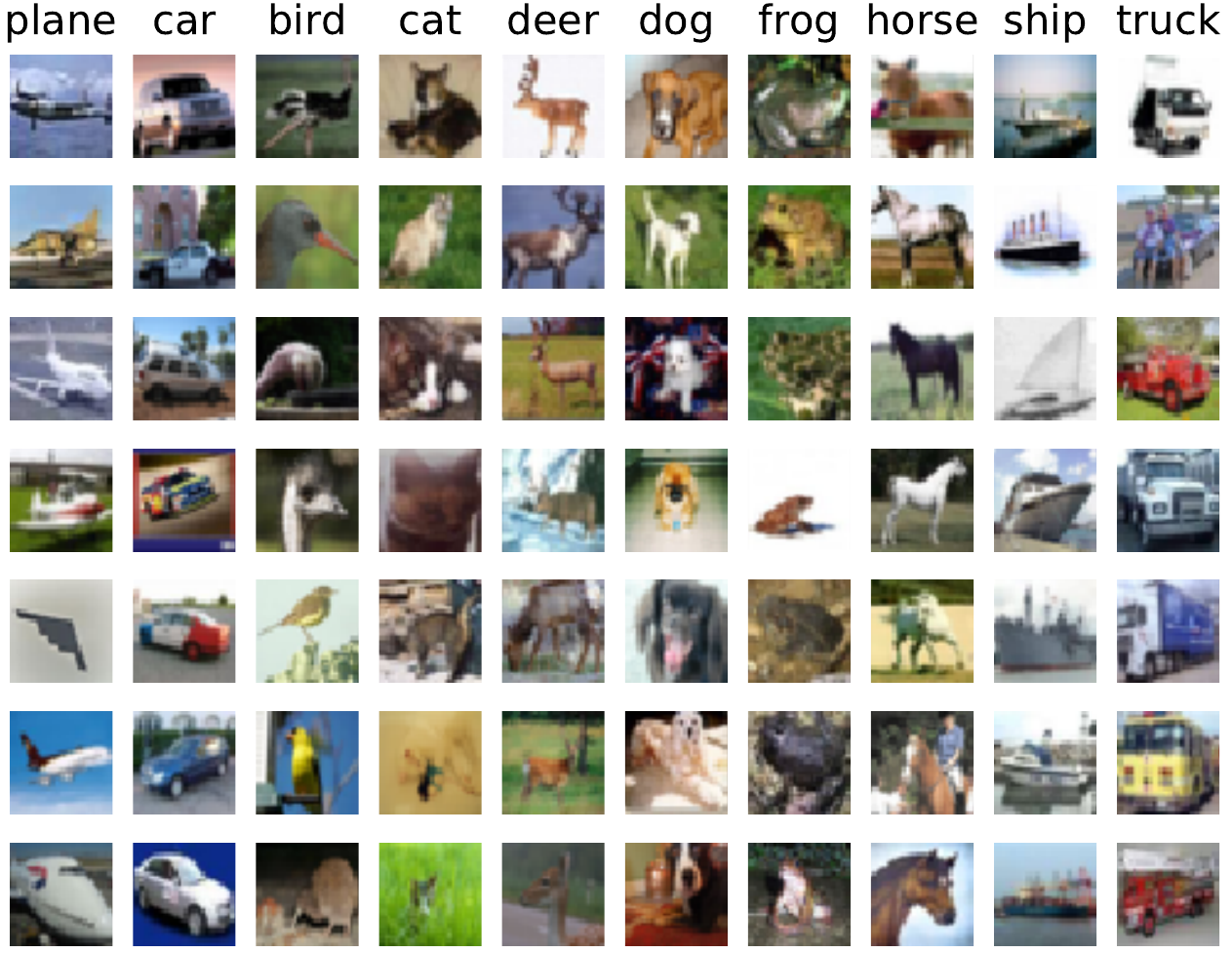} &
\includegraphics[scale=.34]{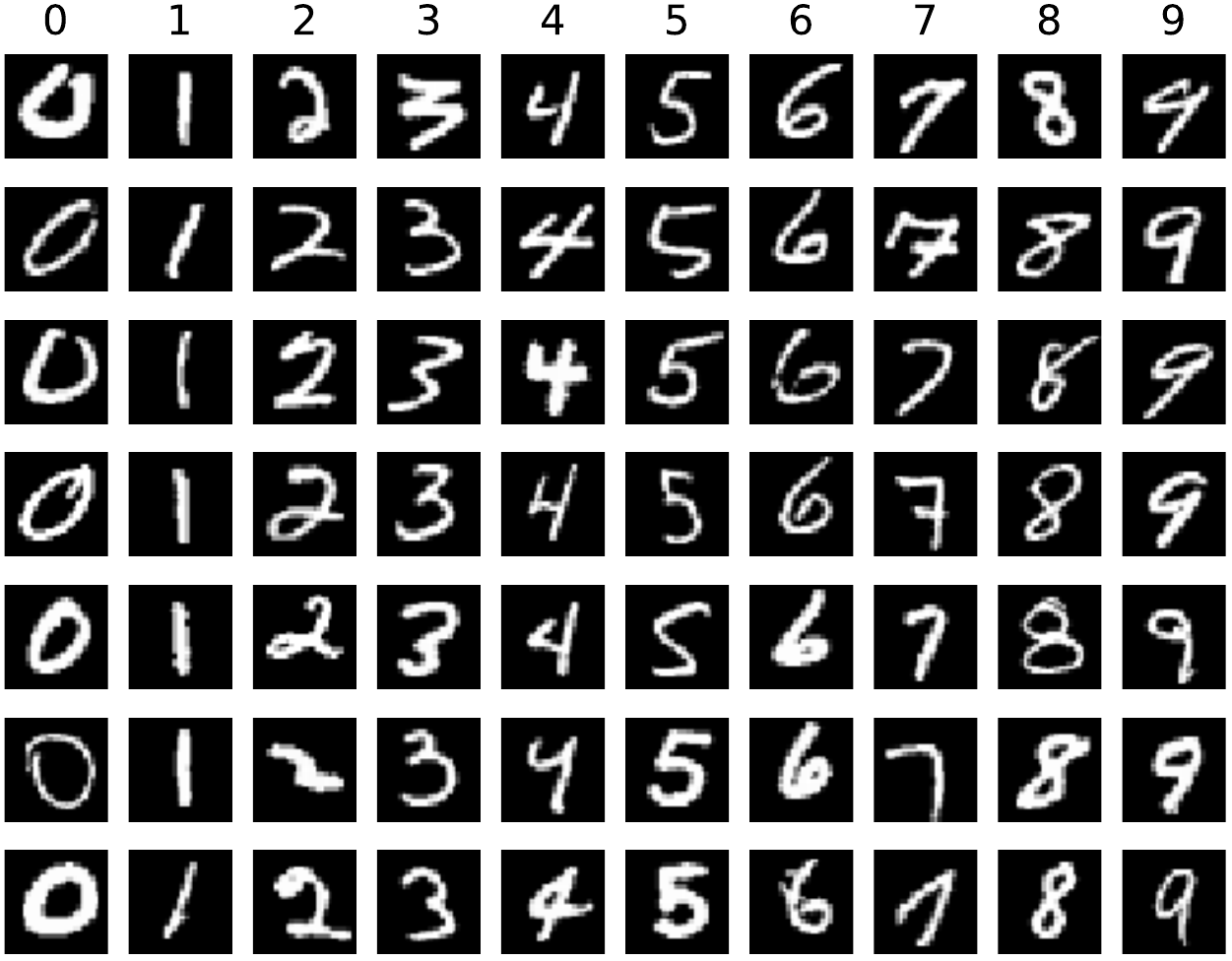}
\end{tabular}
\caption{Sample images from CIFAR10 (left) and MNIST (right).}
\label{Fig:CIFAR10MNIST}
\end{figure}

\medskip
\noindent{\textbf{Datasets.}}\quad We demonstrate and analyze the performance of the proposed AGCN model on three citation networks (Cora, Citseer, and Pubmed) and two image datasets (MNIST and CIFAR10). The summary descriptions of these benchmark datasets are as follows:
\begin{itemize}
\item Cora is a citation network dataset consisting of 2,708 nodes representing scientific publications and 5,429 edges representing citation links between publications. All publications are classified into 7 classes (research topics). Each node is described by a binary feature vector indicating the absence/presence of the corresponding word from the dictionary, which consists of 1,433 unique words.
\item Citeseer is a citation network dataset composed of 3,312 nodes representing scientific publications and 4,723 edges representing citation links between publications. All publications are classified into 6 classes (research topics). Each node is described by a binary feature vector indicating the absence/presence of the corresponding word from the dictionary, which consists of 3,703 unique words.
\item Pubmed is a citation network dataset containing 19,717 scientific publications pertaining to diabetes and 44,338 edges representing citation links between publications. All publications are classified into 3 classes. Each node is described by a TF/IDF weighted word vector from the dictionary, which consists of 500 unique words.
\item CIFAR10 is an image dataset consisting of 60,000 natural color images, each of which is $32\times 32\times 3$ in size and has three color channels (RGB), as shown in Figure~\ref{Fig:CIFAR10MNIST} (left). All images in the dataset are classified into 10 classes, with 6,000 images per class. There are 50,000 training images and 10,000 testing images. In our experiments, we randomly select 10,000 images (1,000 images per class) to perform evaluation. For each image, a convolutional neural network is used to extract a feature descriptor, followed by graph construction using the $k$-nearest neighbors algorithm with $k=8$.
\item MNIST is an image dataset consisting of 70,000 grayscale images of handwritten digits from 0 to 9 (i.e. 10 classes), as shown in Figure~\ref{Fig:CIFAR10MNIST} (right). There are 60,000 training images and 10,000 testing images taken from American Census Bureau employees and American high school students, respectively. We randomly select 1,000 images from each digit for evaluation. Each image is $28\times 28$ in size, and hence it is represented as a 784-dimensional feature vector. The $k$-nearest neighbors algorithm with $k=8$ is used to construct the graph for each handwritten digit.
\end{itemize}

\medskip
\noindent\textbf{Baseline methods.}\quad We evaluate the performance of AGCN against several graph-based feature learning models, including DeepWalk~\cite{Perozzi:14}, Chebyshev networks (ChebyNet)~\cite{Defferrard:16}, GCN~\cite{Kipf:17}, mixture model network (MoNet)~\cite{Monti:17}, graph attention network (GAT)~\cite{Velickovic:17}, jumping knowledge network (JK-Net)~\cite{Keyulu:18}, deep graph infomax (DGI)~\cite{Velickovic:19}, graph wavelet neural network (GWNN)~\cite{Bingbing:19}, Lanczos network (LanczosNet)~\cite{Liao:19}, graph isomorphism network (GIN)~\cite{KXu:19}, simple graph convolution (SGC)~\cite{FWu:19}, and GCN with pair normalization (GCN-PN)~\cite{Lingxiao:20}. For baselines, we mainly consider methods that are closely related to AGCN and/or the ones that are state-of-the-art node classification frameworks. A brief description of these standard baselines can be summarized as follows:
\begin{itemize}
\item DeepWalk is a deep learning based framework that learns latent representations of nodes in a graph by leveraging local information obtained from truncated random walks.
\item ChebyNet is an efficient spectral-domain graph convolutional neural network that uses recursive Chebyshev polynomial spectral filters to avoid explicit computation of the Laplacian eigenvectors.
\item GCN is a semi-supervised graph-based deep learning framework that uses an efficient layer-wise propagation rule that is based on a first-order approximation of spectral graph convolutions.
\item MoNet is a spatial-domain graph convolutional neural network that employs a mixture of Gaussian kernels with learnable parameters to model the weight function of pseudo-coordinates, which are associated to the neighboring nodes of each graph node.
\item GAT is graph-based neural network architecture that uses an attention mechanism to assign self-attention scores to neighboring node embeddings.
\item JK-Net is a graph representation learning approach that learns to selectively exploit information from neighborhoods of differing locality and combines different aggregations at the last layer.
\item DGI is graph representation learning framework, which leverages local mutual information maximization across the graph's patch representations to learn node embeddings in an unsupervised manner.
\item GWNN is a spectral convolutional neural network, which uses spectral graph wavelets for node feature aggregation.
\item LanczosNet is a multiscale graph convolutional network for learning node embedding, which leverages the Lanczos algorithm to construct a low rank approximation of the graph Laplacian for graph convolution.
\item GIN is a simple graph neural network architecture, which maps isomorphic graphs to the same representation and non-isomorphic ones to different representations. It is proved to be as powerful as the Weisfeiler-Lehman test for graph isomorphism.
\item SGC is a simplified GCN architecture, which consists of a linear feature propagation scheme, followed by multi-class logistic regression.
\item GCN-PN is a graph convolutional network, which uses a pair normalization layer to help prevent oversmoothing in deeper graph neural networks.
\end{itemize}
We also compare our approach with multi-layer perceptron (MLP), manifold regularization (ManiReg)~\cite{Belkin:06}, semi-supervised embedding (SemiEmb)~\cite{Weston:12}, LP~\cite{XWu:12}, iterative classification algorithm (ICA)~\cite{PSen:08}, and Planetoid~\cite{Yang:16}.

\begin{figure}[!htb]
\centering
\includegraphics[scale=.67]{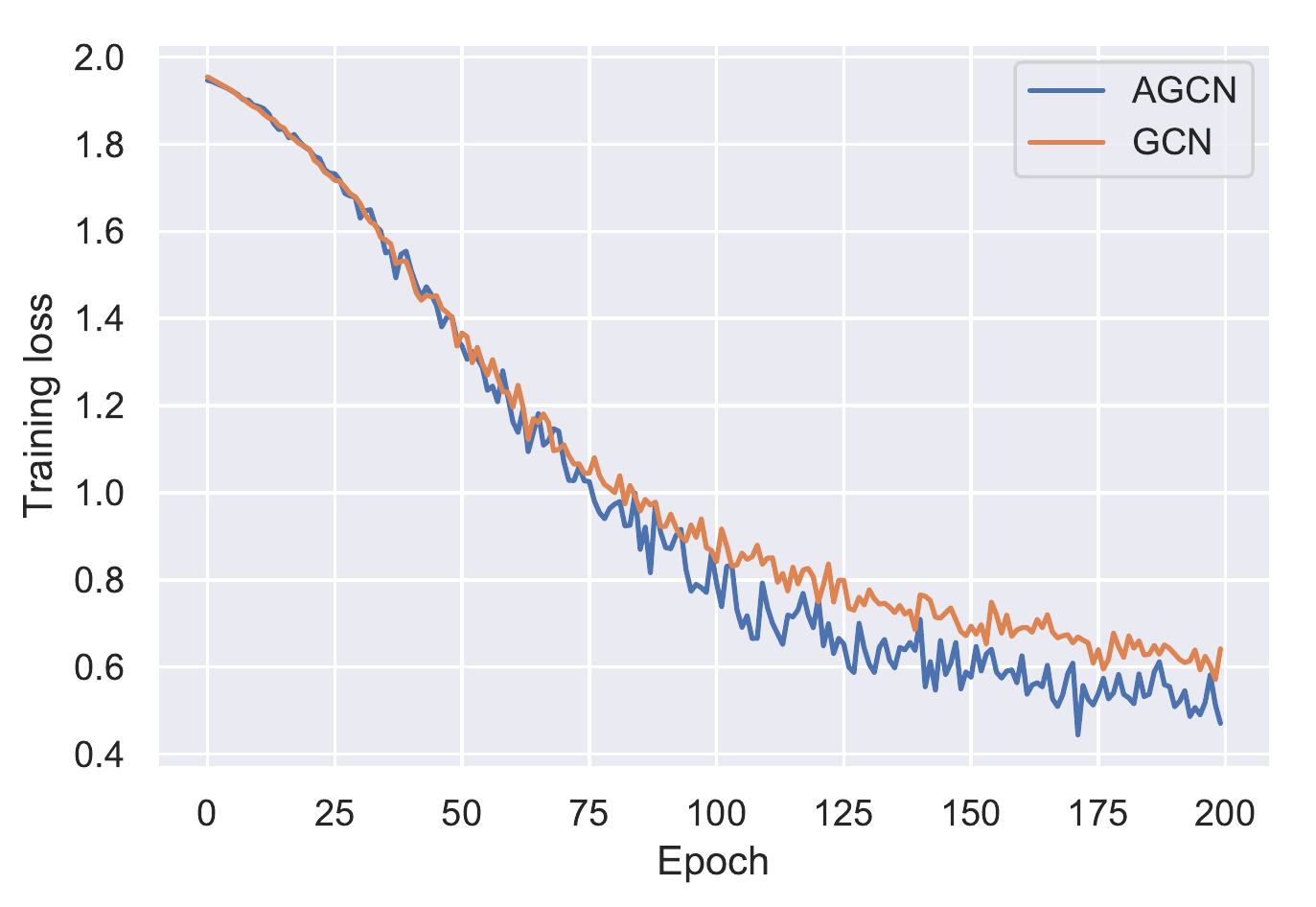}
\caption{Model training history comparison between GCN and proposed AGCN model on the Cora dataset.}
\label{Fig:Loss_curve}
\end{figure}

\medskip
\noindent{\textbf{Implementation details.}}\quad For fair comparison with prior work, we follow the same experimental setup as~\cite{Kipf:17}. For the CIFAR10 and MNIST image datasets, we randomly select 3,000 images as labeled samples and used the remaining images as unlabeled samples. For unlabeled samples, we select 1,000 images for validation and used the remaining 6,000 images as test samples. All the reported accuracy results of AGCN are averaged over 10 runs with different splits for training, validation and test sets. We train the proposed AGCN model for 200 epochs using Adam optimizer~\cite{Kingma:15} with learning rate 0.01. The training is stopped when the validation loss does not decrease after 10 consecutive epochs. The values of the cross-entropy metric are recorded at the end of each epoch on the training set. The performance comparison between AGCN and GCN over training epochs on the training set is illustrated in Figure~\ref{Fig:Loss_curve}, which shows that the proposed AGCN model yields lower training loss values, indicating higher predictive accuracy. The value of the hyperparameter $\beta$ is optimized using grid search over the set $\{0, 0.1, 0.2,\dots,5\}$.

\subsection{Results}
The performance of our model is evaluated by conducting a comprehensive comparison with standard baseline methods for node classification using average accuracy as an evaluation metric. The average classification accuracy results in percent are summarized in Table~\ref{Tab:class}. Results for baseline methods on the citation networks are taken from the GAT paper~\cite{Velickovic:17}, and from the corresponding baseline papers for the image datasets. As shown in Table~\ref{Tab:class}, our AGCN model outperforms GCN on the citation networks as well as on the image datasets. While DeepWalk does well on the MNIST dataset, it performs poorly on the citation networks compared to AGCN. In addition, AGCN outperforms GAT on the image datasets and the Pubmed citation network, and performs on par with GAT on the Cora dataset. The average accuracy of AGCN on the CIFAR10 dataset is 70\%, indicating a performance improvement of 3.4\% over GAT. Similarly, AGCN performs on par with DGI on Citeseer, but yields better performance on Cora and Pubmed. Also, AGCN achieves better performance than GWNN, a spectral approach that requires computing an eigendecomposition of the Laplacian, which is usually time- and space-consuming. We can see that AGCN outperforms GIN and LanczosNet across all the citation network datasets, with performance gains of 5.4\% and 3.5\% on Cora and 5.4\% and 3.5\% on Citeseer, respectively. Moreover, AGCN performs better than SGC and JK-Net on Cora and Pubmed. Interestingly, despite its simplicity, AGCN achieves a higher accuracy than GCN-NP with improvements of 5.8\% and 4\% on Citeseer and Cora, respectively. These results demonstrate the significant prediction ability of AGCN in semi-supervised node classification. It is important to mention that both JK-Net and GCN-PN use additional steps such as skip connections and normalization layers to tackle the issue of oversmoothing on graph neural networks, while our proposed AGCN model integrates an oversmoothing prevention term into its neighborhood aggregation scheme using a single step. In other words, the proposed AGCN network is trained in an end-to-end fashion, and eliminates the need to augment deep GNN models with normalization layers or by inserting residual/skip connections between the network's layers in order to improve performance.

\begin{table}[!htb]
\setlength{\tabcolsep}{.47em}
\caption{Classification accuracy results on three citations networks and two image datasets. Boldface numbers indicate the best classification performance.}
\medskip
\centering
\begin{tabular}{l*{5}{c}}
\toprule
& \multicolumn{5}{c}{Average accuracy (\%)}\\
\cmidrule(lr){2-6}
Method &  Cora & Citseer & Pubmed & MNIST & CIFAR10\\
\midrule
MLP &55.1& 46.5 &71.4&$-$&$-$\\
ManiReg & 59.5& 60.1& 70.7&94.6&59.7\\
SemiEmb & 59.0 &59.6 &71.7&$-$&$-$\\
LP & 68.0 &45.3& 63.0&83.4&60.4\\
DeepWalk &67.2& 43.2 &65.3&\textbf{95.3}&61.3\\
ICA & 75.1 &69.1& 73.9&$-$&$-$\\
Planetoid & 75.7& 64.7& 77.2&$-$&$-$\\
ChebyNet & 81.2 &69.8& 74.4&$-$&$-$\\
GCN &81.5& 70.3& 79.0&91.0&61.0\\
MoNet &81.7 &$-$ & 78.8 &$-$&$-$\\
GAT & \textbf{83.0} & 72.5 & 79.0 & 92.8 & 66.6\\
DGI & 82.3 & 71.8 & 76.8 & $-$ & $-$\\
GWNN & 82.8 & 71.7 & 79.1 & $-$ & $-$\\
LanczosNet & 79.5 & 66.2 & 78.3 & $-$ & $-$\\
GIN & 77.6 & 66.1 & 77.0 & $-$ & $-$\\
SGC & 81.0 & 71.9 & 78.9 & $-$ & $-$\\
JK-Net & 82.7 & \textbf{73.0} & 77.9 & $-$ & $-$\\
GCN-PN & 79.0 & 66.0 & 78.0 & $-$ & $-$\\
\midrule
AGCN &\textbf{83.0}& 71.8   & \textbf{79.5}  &  93.1 &\textbf{70.0} \\
\bottomrule
\end{tabular}
\label{Tab:class}
\end{table}

\begin{figure*}[!t]
\centering
\includegraphics[scale=.9]{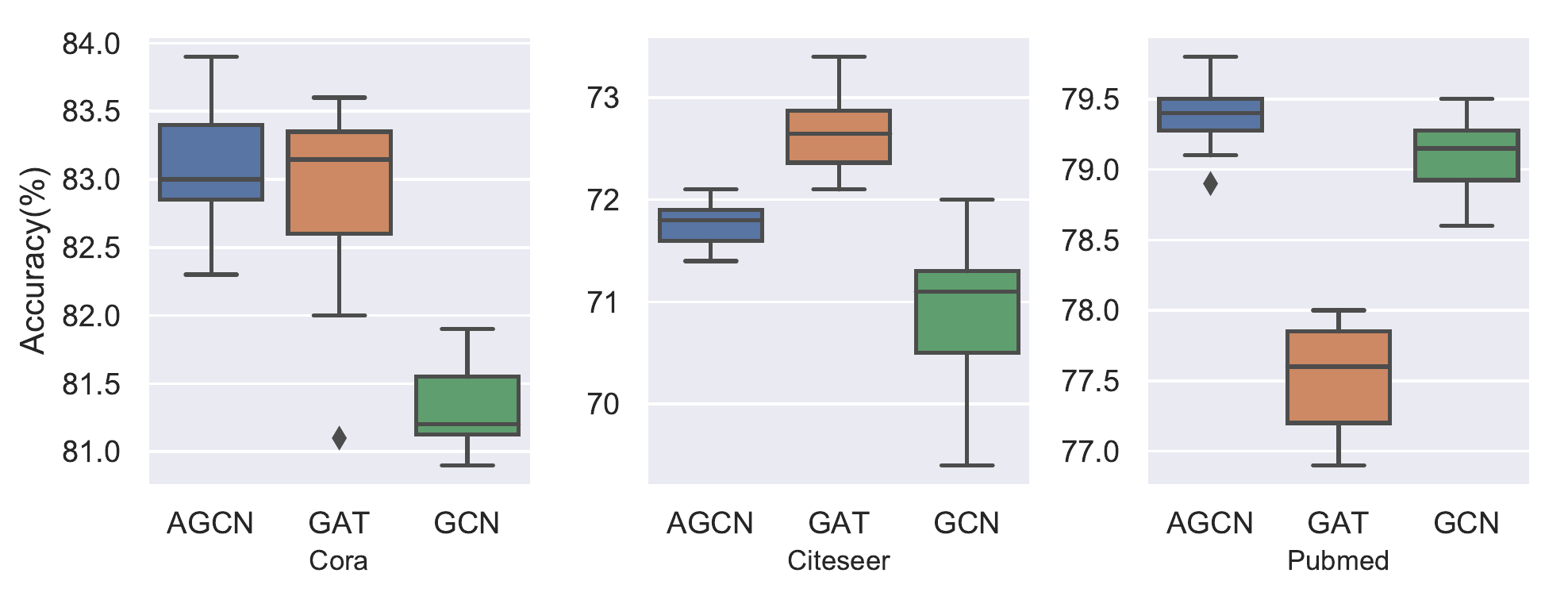}
\caption{Accuracy distributions of AGCN, GAT and GCN on the Cora, Citeseer and Pubmed citation networks.}
\label{Fig:boxplot}
\end{figure*}

Using box plots, Figure~\ref{Fig:boxplot} displays the visual differences in terms of accuracy among AGCN, GAT and GCN on the Cora, Citeseer and Pubmed citation networks. As can be seen, the distribution of the AGCN model has less variability than GCN and GAT on document classification tasks. For instance, the median accuracy score for AGCN on the Pubmed dataset indicates a significant difference in performance between AGCN and the two baseline methods. In addition, the box for AGCN is short, meaning that the accuracy values consistently hover around the average accuracy. However, the box for GAT is taller, implying variable accuracy values compared to AGCN.

\medskip
\noindent\textbf{Co-training and self-training results.}\quad Using the co-training and self-training approaches~\cite{Li:18}, we compare AGCN with GCN on the Cora, Citeseer and Pubmed datasets with label rates (i.e. proportion of labeled nodes that are used for training) 0.036, 0.052, and 0.003, respectively.

\begin{figure}[!htb]
\centering
\includegraphics[scale=.75]{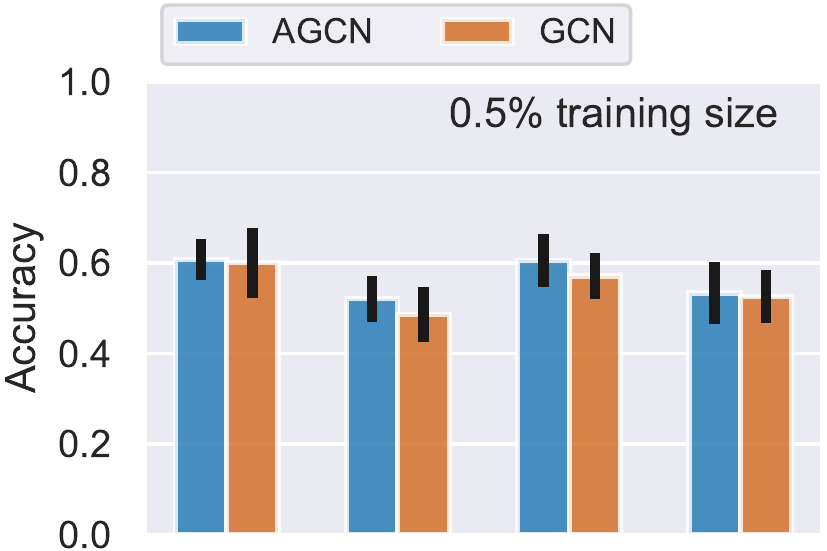}\\
\includegraphics[scale=.75]{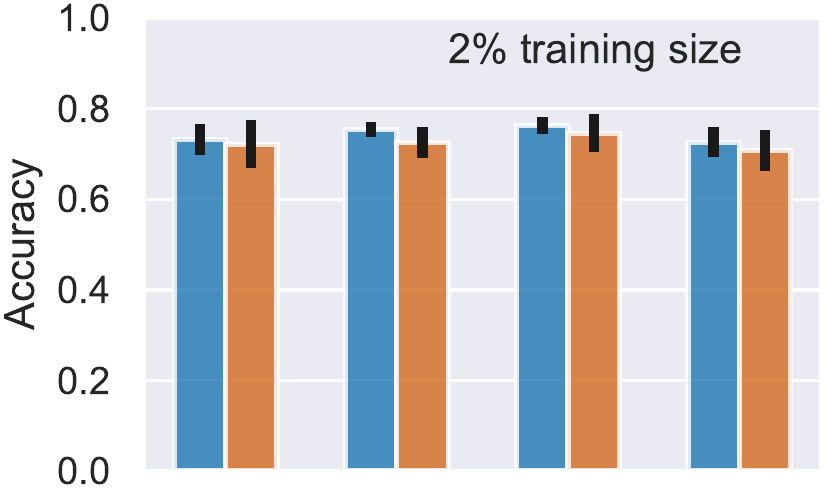}\\
\includegraphics[scale=.75]{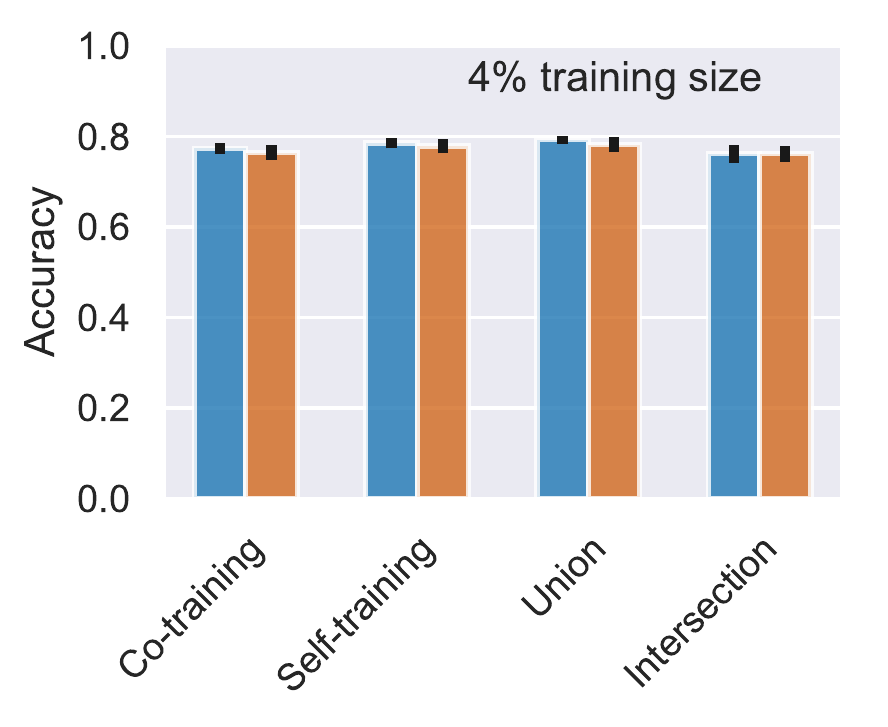}
\caption{Classification accuracy of AGCN compared to GCN for different training set sizes on the Cora dataset.}
\label{fig:cora}
\end{figure}

\begin{figure}[!htb]
\centering
\includegraphics[scale=.75]{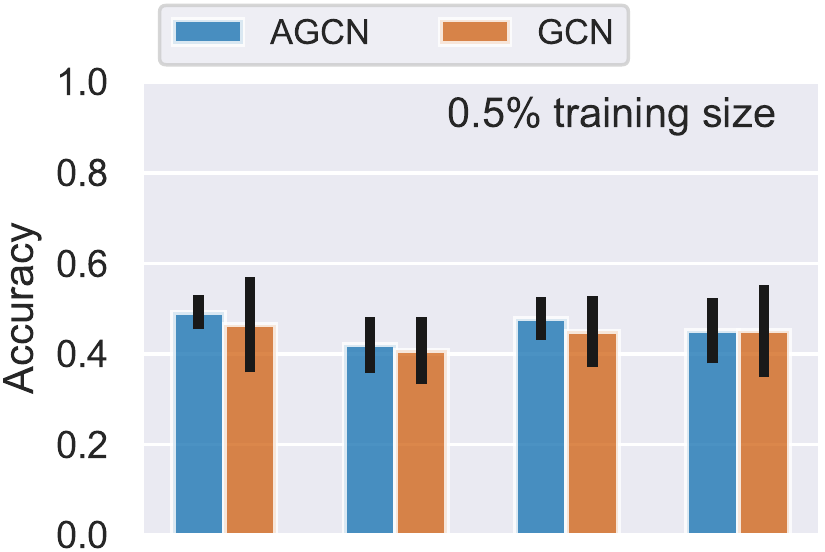}\\
\includegraphics[scale=.75]{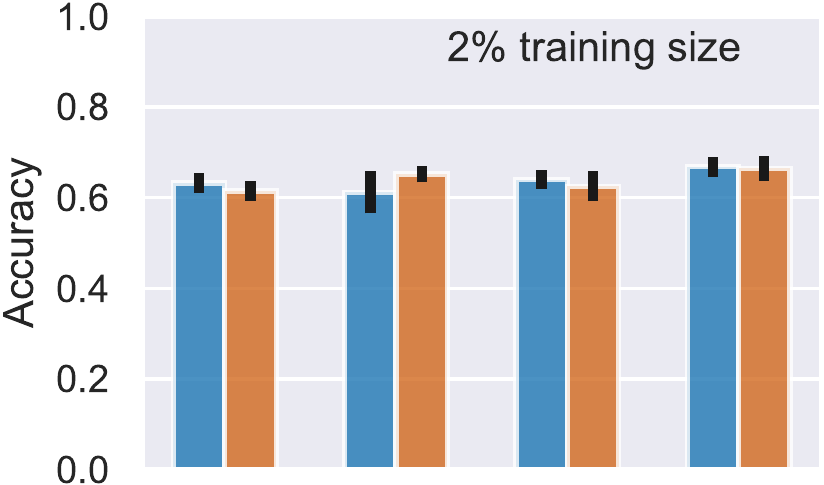}\\
\includegraphics[scale=.75]{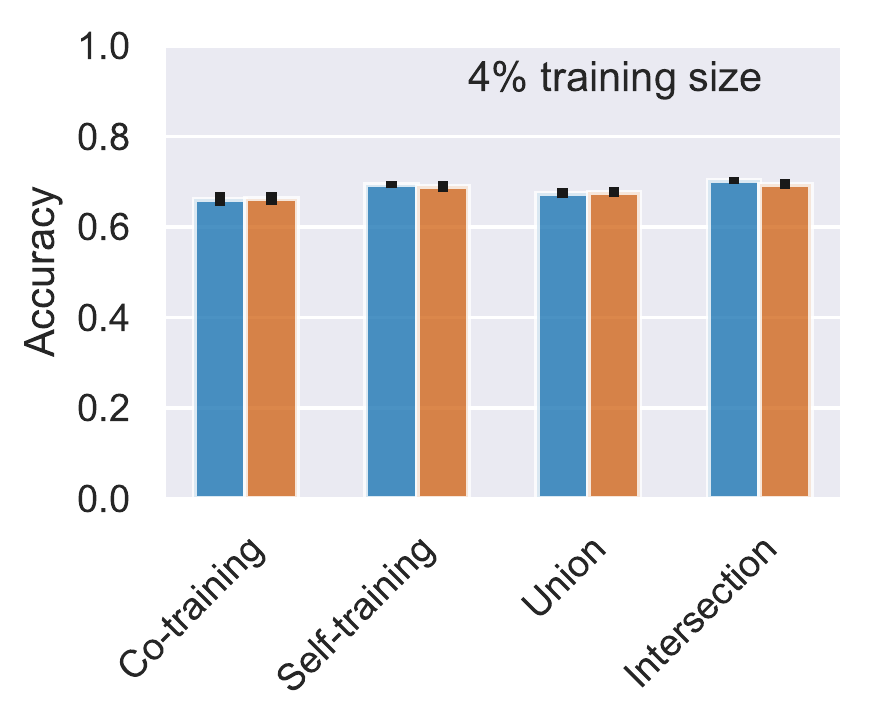}
\caption{Classification accuracy of AGCN compared to GCN for different training set sizes on the Citeseer dataset.}
\label{fig:citeSeer}
\end{figure}

\begin{figure}[!htb]
\centering
\includegraphics[scale=.75]{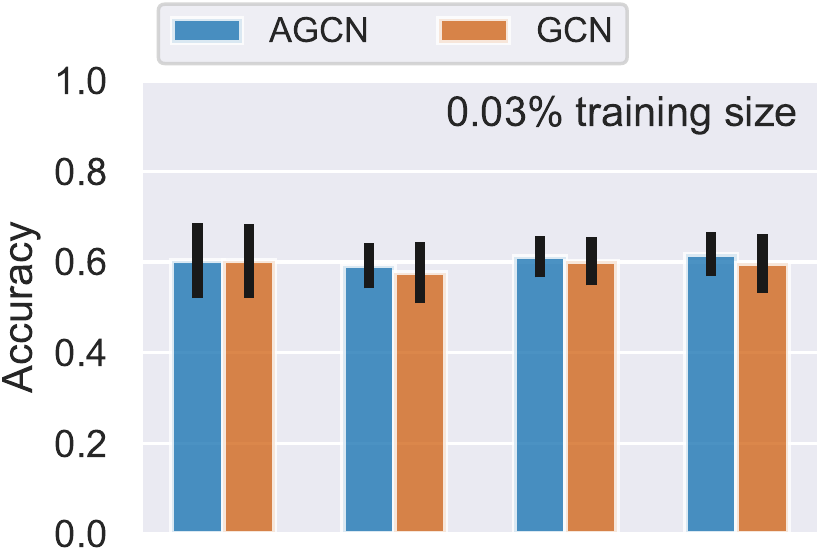}\\
\includegraphics[scale=.75]{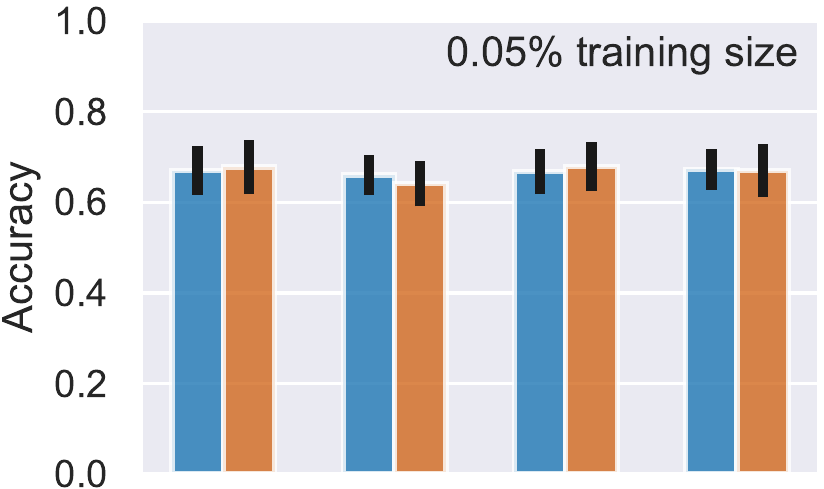}\\
\includegraphics[scale=.75]{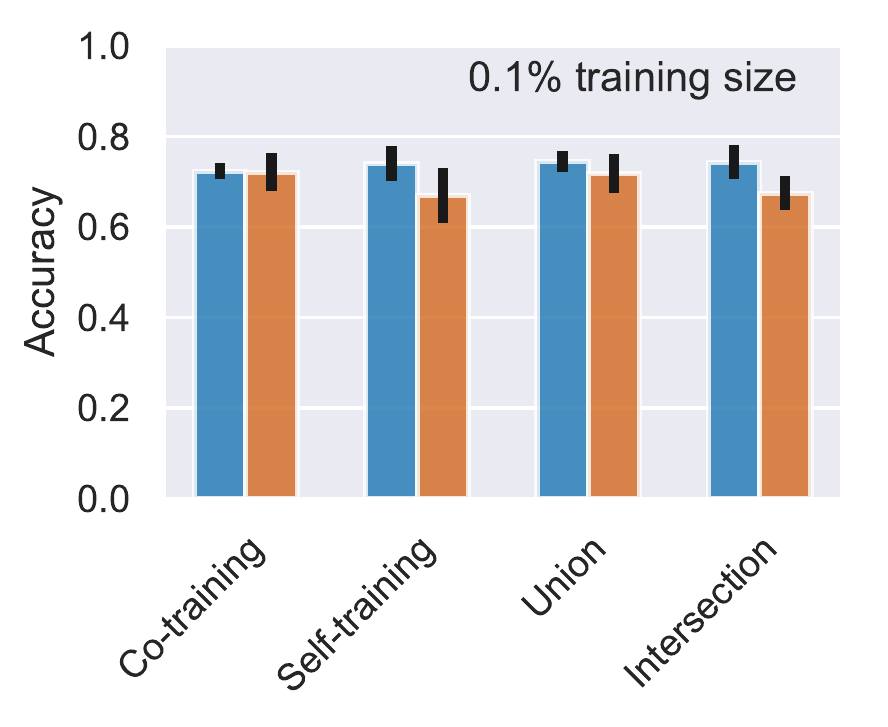}
\caption{Classification accuracy of AGCN compared to GCN for different training set sizes on the Pubmed dataset.}
\label{fig:pubmed}
\end{figure}

We apply co-training and self-training approaches as well as their intersection and union to train our AGCN model and compare it to GCN. The accuracy results for these four approaches are reported in Figures~\ref{fig:cora}, \ref{fig:citeSeer} and \ref{fig:pubmed} on the Cora, Citeseer and Pubmed citation networks, respectively, using training rates of $0.5\%$, $2\%$ and $4\%$ for Cora and Citseer, and $0.03\%$, $0.05\%$ and $0.1\%$ for Pubmed. In each bar plot, the bars display the mean and standard error accuracy over 10 runs for both AGCN and GCN using co-training, self-training, union and intersection. In co-training, a partially absorbing random walk is used to find the confidence of node $i$ belongs to class $c$. The most confident nodes are then added to the training set with label $c$ to train the AGCN model. In self-training, AGCN is applied to find the most confident nodes based on the softmax scores  $\hat{\bm{Y}}\in \mathbb{R}^{N\times C}$ given by Eq.~\eqref{Eq:twolayer}. Then, the most confident nodes are added to the labeled set. On the other hand, union and intersection are a combination of co-training and self-training. Union expands the label set with the most confident predictions obtained by random walk and those obtained by AGCN. As can be seen in these figures, our AGCN framework outperforms GCN in most of the cases, particularly for small training sizes. Moreover, notice that the standard deviations are much smaller than the accuracy improvements, indicating that AGCN is robust to random selection of training and test data. Overall, AGCN is consistently the best performing method, delivering robust classification accuracy results.

\subsection{Statistical Significance Analysis}
In this subsection, we conduct statistical significance tests to compare GCN, GAT and AGCN with the objective of selecting the best performing model. More precisely, we apply one-way analysis of variance (ANOVA) to verify whether there is a statistical difference between their mean accuracy scores. ANOVA tests the hypothesis that all group means are equal versus the alternative hypothesis that at least one group is different from the others:
\begin{equation}
\begin{split}
H_0 &: \mu_1=\mu_2=\mu_3 \\
H_1 &: \text{not all group means are equal}
\end{split}
\end{equation}
where $\mu_1,\mu_2,\mu_3$ denote the population means for GCN, GAT and AGCN, respectively. While ANOVA is based on the assumption that all sample populations are normally distributed, it is, however, known to be robust to modest violations of the normality assumption.

We perform one-way ANOVA for the accuracy scores data obtained by GCN, GAT and AGCN on the Cora, Citeseer and Pubmed datasets. These results correspond to 10 runs with different splits for training, validation and test sets. As shown in Table~\ref{Tab:p-value}, the small $p$-values ($<0.05$) indicate that differences between accuracy means are statistically significant, where $\alpha=0.05$ is the significance level. A significance level of 0.05 indicates a $5\%$ risk of concluding that a difference exists when there is no actual difference.

\begin{table}[!htb]
\caption{One-way ANOVA $p$-values for the accuracy scores data obtained by AGCN, GCN and GAT on the Cora, Citeseer and Pubmed datasets.}
\medskip
\centering
\begin{tabular}{l*{3}{c}}
\cmidrule(lr){2-4}
 &  Cora & Citeseer & Pubmed \\
\midrule
$p$-value &1.0$\times 10^{-3}$& 2.51$\times 10^{-7}$ &1.21$\times 10^{-6}$ \\
\bottomrule
\end{tabular}
\label{Tab:p-value}
\end{table}

\begin{figure}[!htb]
\includegraphics[scale=.4]{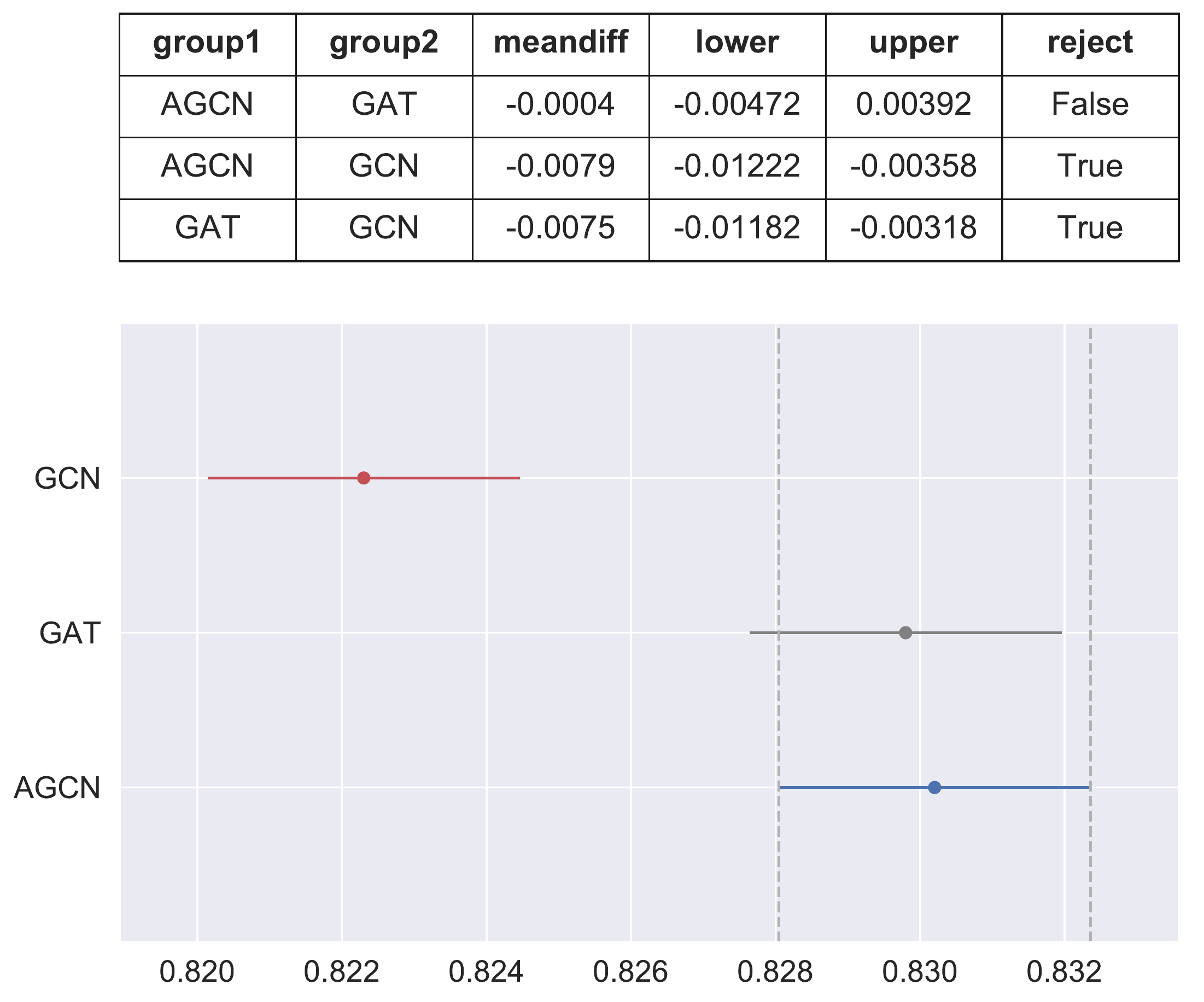}
\caption{Pairwise multiple comparison between AGCN, GAT, and GCN methods using Tukey's test on the Cora dataset.}
\label{fig:TukeyCora}
\end{figure}

\begin{figure}[!htb]
\includegraphics[scale=.4]{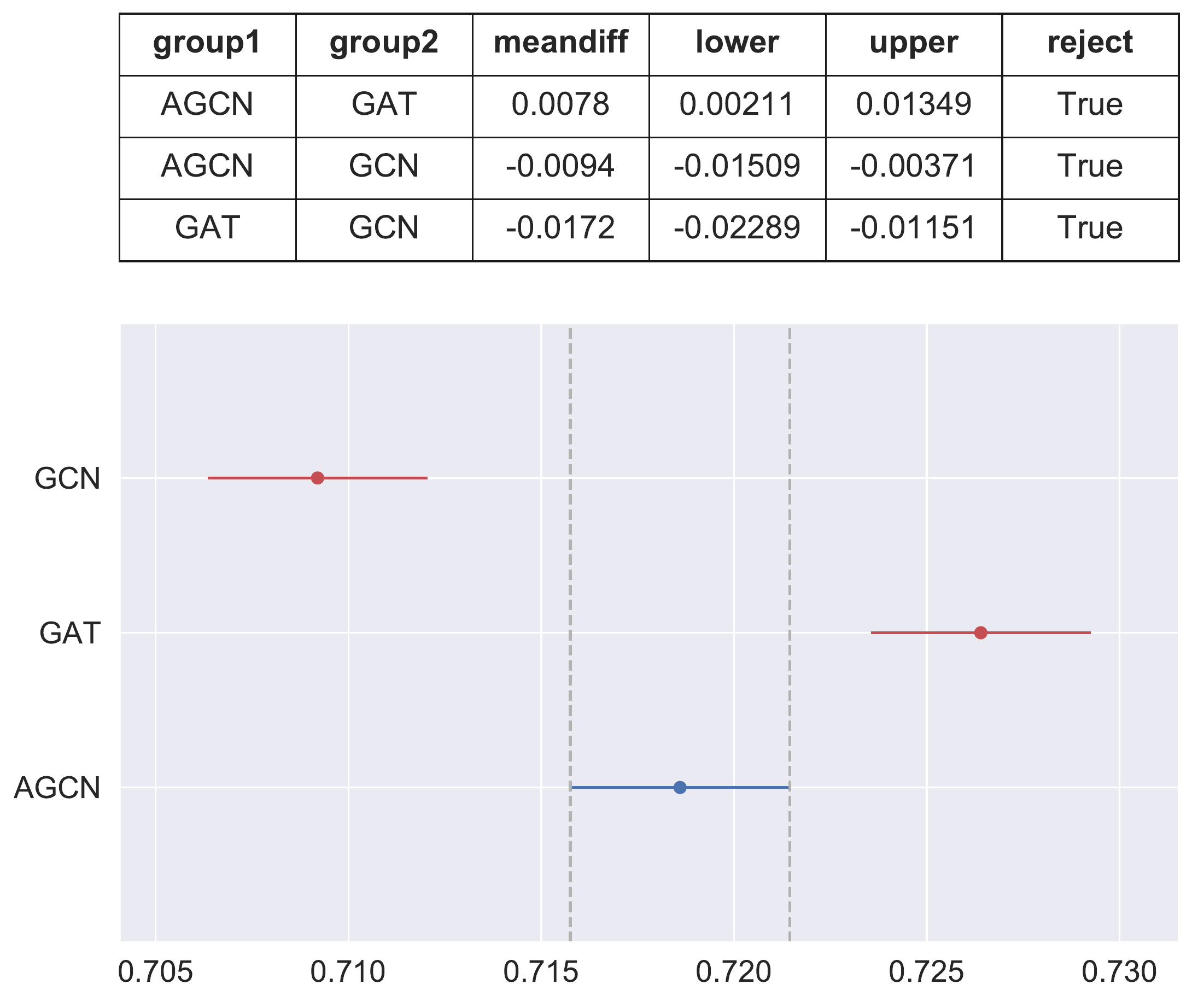}
\caption{Pairwise multiple comparison between AGCN, GAT, and GCN methods using Tukey's test on the Citeseer dataset.}
\label{fig:TukeyCiteseer}
\end{figure}

\begin{figure}[!htb]
\includegraphics[scale=.4]{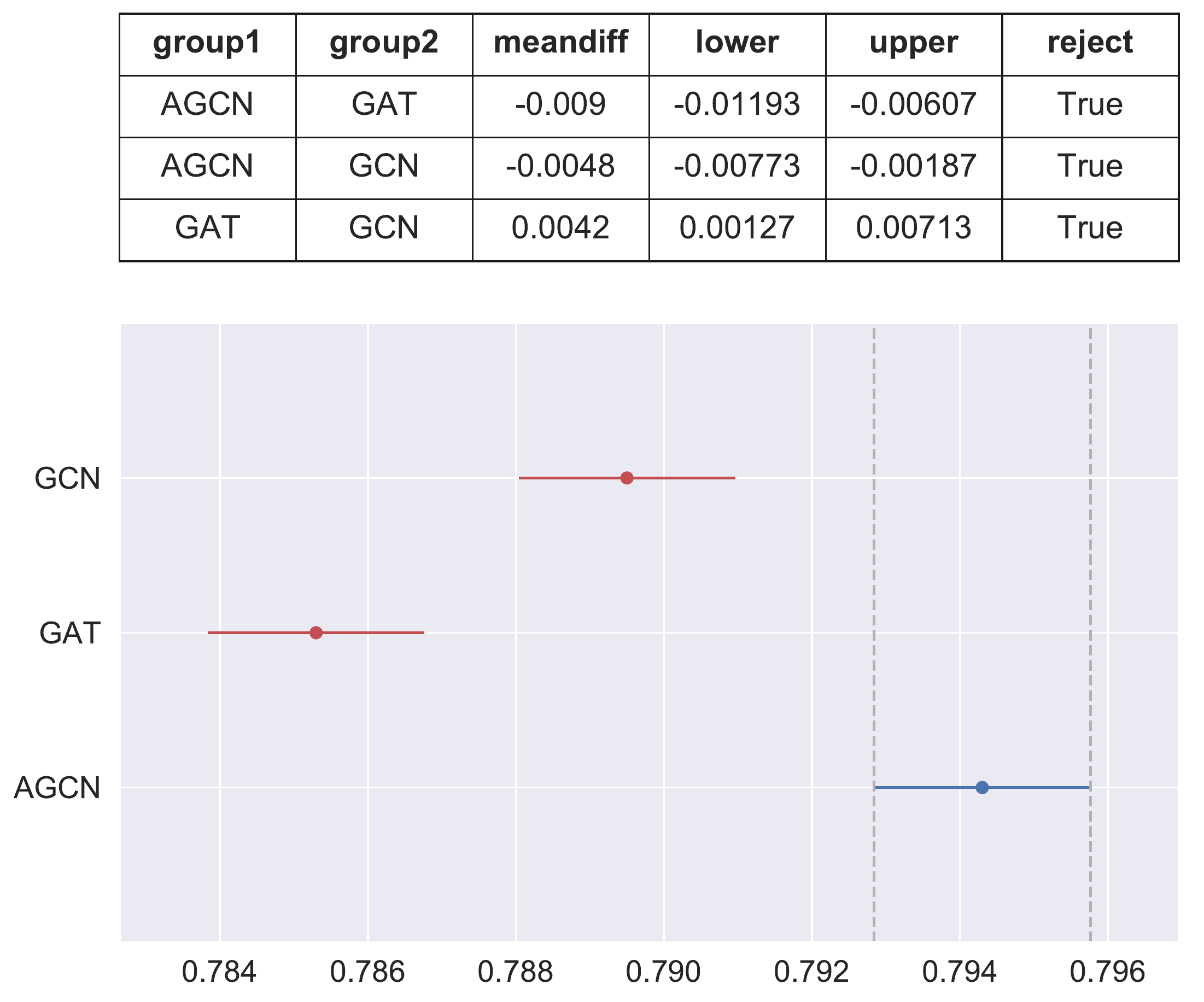}
\caption{Pairwise multiple comparison between AGCN, GAT, and GCN methods using Tukey's test on the Pubmed dataset.}
\label{fig:TukeyPubmed}
\end{figure}

Since our ANOVA analysis shows an overall statistically significant difference in group means, we now need to determine which specific groups (compared with each other) are different in terms of mean accuracies by performing multiple pairwise comparison (post-hoc comparison) analysis using Tukey's test, which compares all possible pairs of means. The pairwise multiple comparison results of GCN, GAT, and AGCN methods using Tukey's test on the Cora, Citeseer and Pubmed datasets are shown in Figures~\ref{fig:TukeyCora}, \ref{fig:TukeyCiteseer} and \ref{fig:TukeyPubmed}, respectively. The table above each figure reports the results of the multiple comparison of means. In the case of the Cora dataset, for instance, we can see from the result shown in the table of Figure~\ref{fig:TukeyCora} that we reject the hypotheses that ``AGCN and GCN'' and ``GAT and GCN'' have the same mean, but we fail to reject the ``AGCN and GAT'' pair and conclude that they have equal mean accuracies. In the meandiff column, the difference of accuracy means between related groups is reported, followed by lower and upper limits for 95\% confidence intervals for the true mean difference. In the reject column, ``True'' indicates that there is significant evidence to reject the null hypothesis at the given significance level, i.e. there is a significant statistical difference between the means. The results from Tukey's test reported in the tables of Figures~\ref{fig:TukeyCiteseer} and \ref{fig:TukeyPubmed} show that we reject the hypotheses that ``AGCN and GAT'', ``AGCN and GCN'' and ``GAT and GCN'' have the same mean, indicating statistical significant differences.

In each figure, the 95$\%$ confidence intervals plots are displayed as horizontal bars. The blue bar in Figure~\ref{fig:TukeyCora} shows the comparison interval for the AGCN mean accuracy, which does not overlap with the comparison interval for the GCN mean accuracy, shown in red. The comparison interval for the GAT mean accuracy, shown in gray, overlaps with the comparison interval for the AGCN mean accuracy. Hence, the accuracy means for AGCN and GAT are not significantly different from each other on the Cora dataset, meaning that both methods performs on par with each other.

In Figure~\ref{fig:TukeyCiteseer}, the blue bar shows the comparison interval for the AGCN mean accuracy, which does not overlap with the comparison intervals for the GCN and GAT mean accuracies, shown in red. The disjoint comparison intervals indicate that the group means are significantly different from each other. Similarly, the blue bar in Figure~\ref{fig:TukeyPubmed} shows the comparison interval for the AGCN mean accuracy, which does not overlap with the comparison intervals for the GCN and GAT mean accuracies, shown in red. Hence, we conclude that AGCN is the best performing method on the Pubmed dataset. This visual observation is consistent with the results reported in Table~\ref{Tab:class}.

\subsection{Visualization}
\begin{figure}[!htb]
\includegraphics[scale=.67]{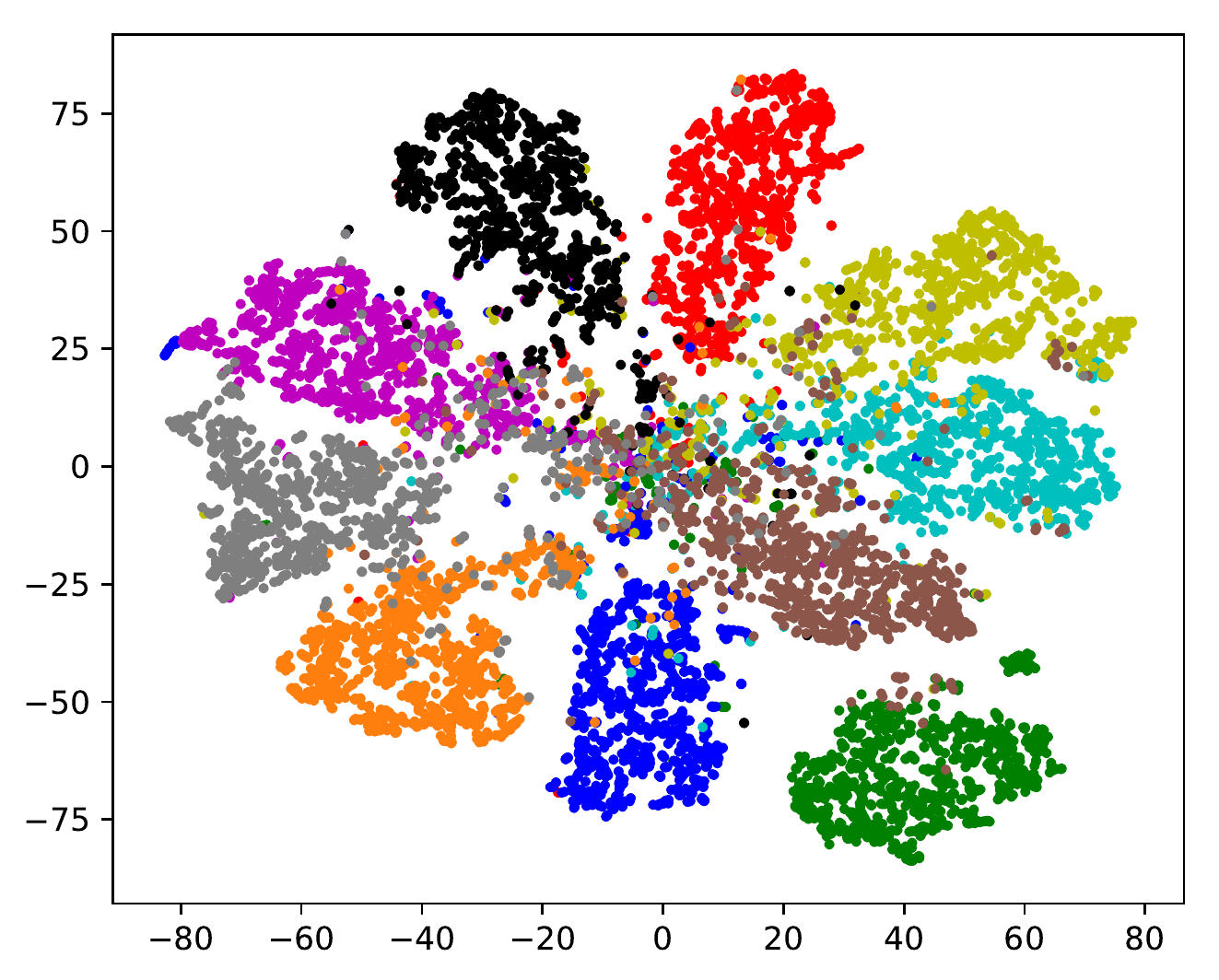}\\[2ex]
\includegraphics[scale=.67]{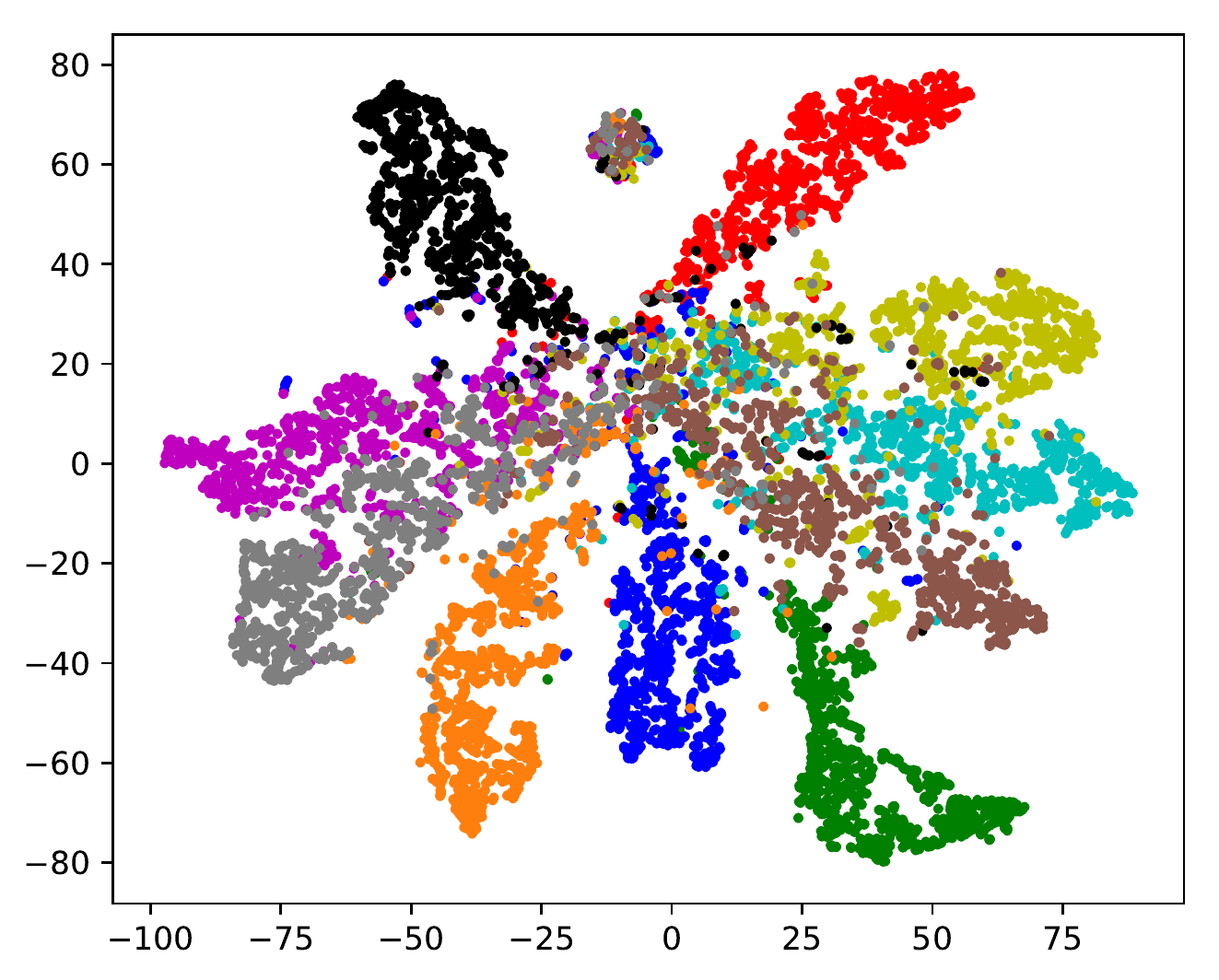}
\caption{t-SNE feature visualization of the output embeddings by the first convolutional layer of AGCN (top) and GCN (bottom), respectively, on the MNIST dataset. Each color denotes a class.}
\label{fig:TSNE}
\end{figure}

The feature embeddings learned by AGCN can be visualized using the t-Distributed Stochastic Neighbor Embedding (t-SNE)~\cite{Maaten:08}, which is a dimensionality reduction technique that is particularly well-suited for embedding high-dimensional data into a two- or three-dimensional space. Figure~\ref{fig:TSNE} displays the t-SNE embeddings of the output embeddings by the first convolutional layer of AGCN (top) and GCN (bottom) on the MNIST dataset. As can be seen, the two-dimensional embeddings corresponding to AGCN are more separable than the ones corresponding to GCN. With GCN features, the points are not discriminated very well, while with AGCN features the points are discriminated much better and clearly show the clusters corresponding to the ten digit labels of the MNIST dataset. Hence, AGNC learns more discriminative features for node classification tasks, indicating the superior performance of anisotropic diffusion over linear diffusion. Moreover, Figure~\ref{fig:TSNE} shows that the AGCN approach is exploratory in nature in the sense that it can discover patterns and meaningful sub-groups in a dataset.

\subsection{Robustness to Oversmoothing}
\begin{figure}[!htb]
\centering
\includegraphics[scale=.66]{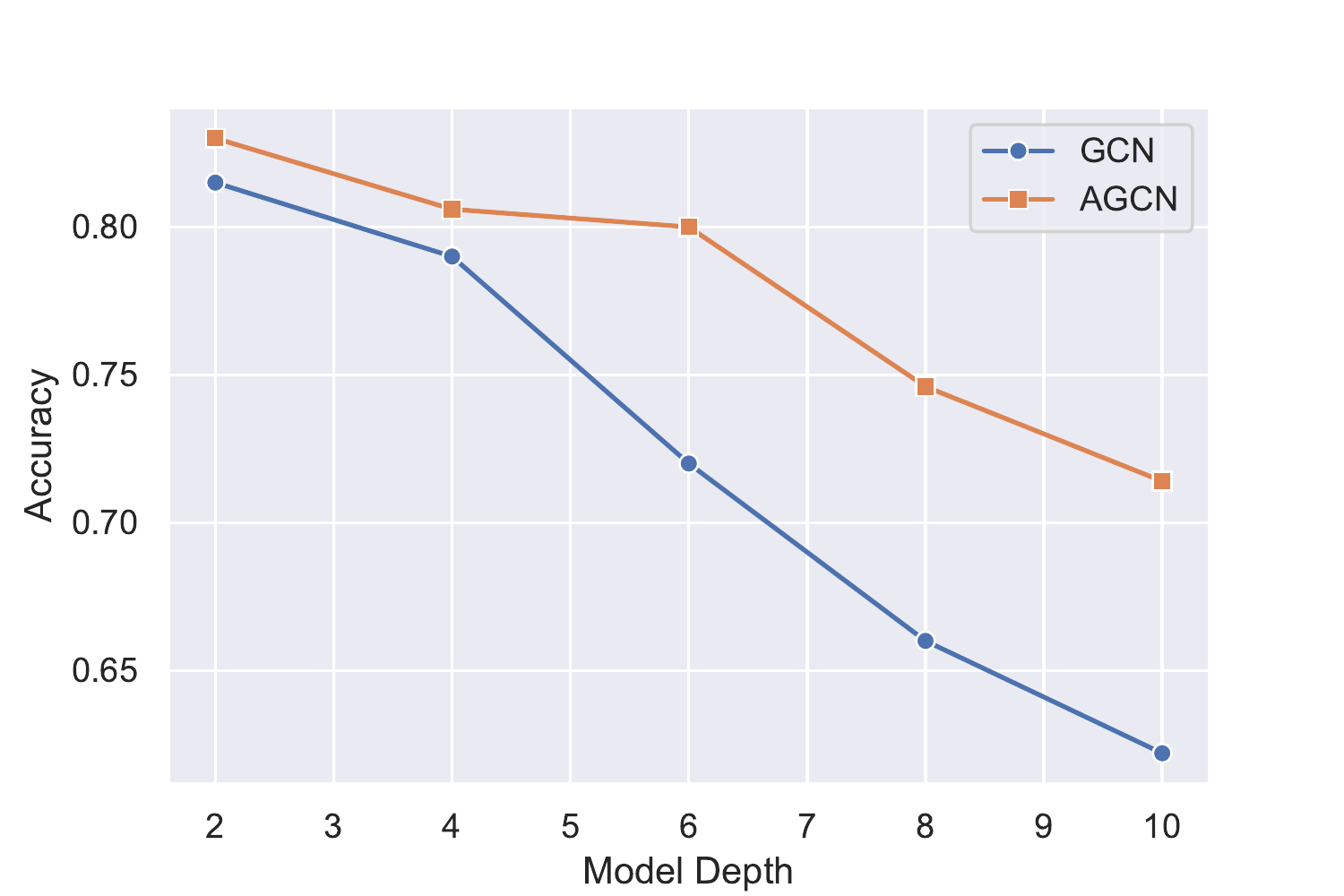}
\caption{Performance comparison between AGCN and GCN on the Cora dataset as we increase the number of layers.}
\label{Fig:oversmooth}
\end{figure}

In order to assess robustness to oversmoothing, we study the performance variation for our multi-layer model on the Cora dataset with respect to the number of layers. Figure~\ref{Fig:oversmooth} shows how the node classification accuracy changes with the network's depth. As can be seen, there is a sharp drop in GCN's accuracy when the number of layers is larger than 4, while AGCN's performance does not significantly degrade as the number of layers increases. Note that the performance gap between AGCN and GCN substantially increases when the network's depth is beyond 4, indicating that AGCN is more robust to oversmoothing. It is worth pointing out that the objective of our proposed anisotropic convolution is to mitigate the oversmoothing issue that causes drops in classification performance for multi-layer GCNs, and not to show that the deeper AGCN, the better. Also, increasing the depth of the network leads to increased number of parameters, causing overfitting and hence resulting in performance drop. As shown in Figure~\ref{Fig:oversmooth}, a 6-layer AGCN yields an accuracy of 80\%, outperforming several baselines including LanczosNet, GIN and GCN-PN. The latter baseline is designed specifically for tackling the issue of oversmoothing in GCNs by employing a normalization layer after each convolutional layer.

\subsection{Parameter Sensitivity Analysis}
\begin{figure}[!htb]
\centering
\includegraphics[scale=.65]{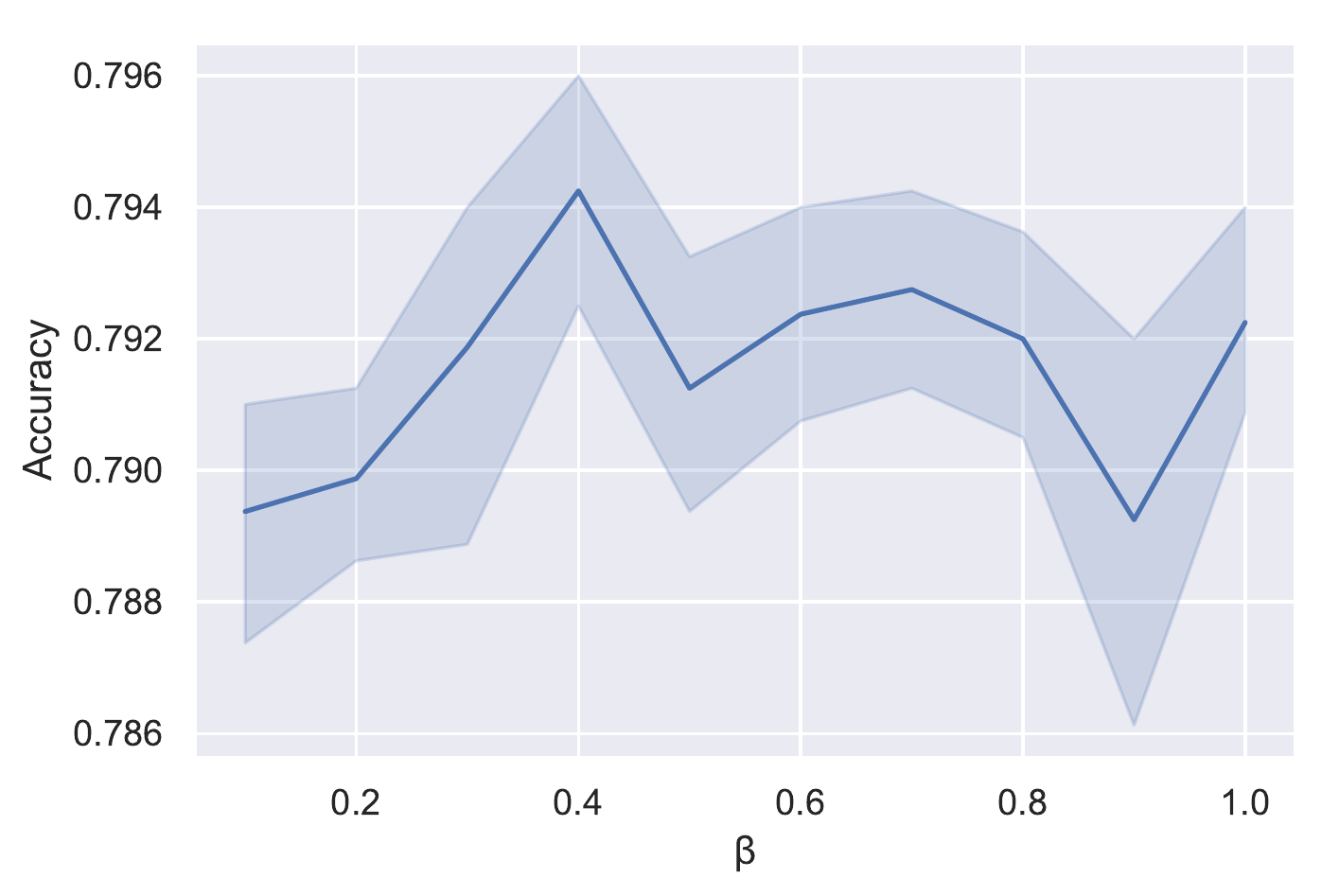}
\caption{AGCN accuracy results for different values of $\beta$ on the Pubmed dataset.}
\label{Fig:beta_parameter}
\end{figure}

For each benchmark dataset used in the experiments, we test the performance of AGCN using different values for the hyper-parameter $\beta$ of the anisotropic diffusion term. In the case of the Pubmed dataset, for instance, the effect of $\beta$ on the performance of AGCN is illustrated in Figure~\ref{Fig:beta_parameter}, which shows the accuracy of AGCN along with the standard error for different values of $\beta$. As can be seen, $\beta=0.4$ yields the best value for the AGCN accuracy on the Pubmed dataset.

\subsection{Discussion}
While the proposed AGCN model shows promising results for end-to-end learning on graphs and mitigates the issue of oversmoothing, its performance is, however, tied to optimizing via grid search with cross-validation the hyper-parameter of the anisotropic diffusion term for each dataset. Another shortcoming of AGCN is that it only takes into account immediate neighbors. This limitation can be circumvented through higher-order message passing by leveraging multi-hop neighbors using powers of the adjacency matrix and hence aggregating learned  node representations from both immediate and distant neighbors.

\section{Conclusion}
In this paper, we introduced an anisotropic graph convolutional network for semi-supervised node classification on graph-structured data by learning efficient representations in an end-to-end fashion. We incorporated a nonlinear smoothness term into the feature diffusion rule of the convolutional neural network in a bid to tackle the issues of oversmoothing and shrinking effect. We demonstrated through extensive experimental results the competitive or superior performance of AGCN in terms of classification accuracy over standard baseline methods on several benchmarks, including citation networks and image datasets. We also showed that AGCN can be integrated into existing graph-based convolutional networks for semi-supervised learning using both co-training and self-training. In addition, we performed a statistical analysis using analysis of variance and pairwise multiple comparison, showing that the performance of our model is better or comparable with the baselines. For future work, we plan to extend the proposed framework to heterogeneous information networks.

\bibliographystyle{ieeetr}
\bibliography{references}

\end{document}